%% file: main.tex
\documentclass[10pt,twocolumn,letterpaper]{article}

\usepackage[pagenumbers]{cvpr} 

\usepackage{color}

\let\oldhat\hat

\renewcommand{\hat}[1]{\oldhat{\mathbf{#1}}}

\input{preamble}

\definecolor{cvprblue}{rgb}{0.21,0.49,0.74}
\usepackage[pagebackref,breaklinks,colorlinks,allcolors=cvprblue]{hyperref}

\def\paperID{******} 
\def\confName{CVPR}
\def\confYear{2026}

\title{SATORI-R1: Incentivizing Multimodal Reasoning through Explicit  \\ Visual Anchoring}

\author{%
  Chuming Shen\textsuperscript{\rm 1}, \quad
  Wei Wei\textsuperscript{\rm 1}\thanks{\quad Corresponding authors.}, \quad 
  Xiaoye Qu\textsuperscript{\rm 1}, \quad
  Yu Cheng\textsuperscript{\rm 2} \\
  \textsuperscript{\rm 1}\ Huazhong University of Science and Technology \\
  \textsuperscript{\rm 2}\ The Chinese University of Hong Kong
\\
    \texttt{\{scm, weiw, xiaoye\}@hust.edu.cn}\quad
\texttt{chengyu@cse.cuhk.edu.hk} \\
}

\begin{document}
\maketitle
\input{sec/0_abstract}    
\input{sec/1_intro}
\input{sec/2_background}
\input{sec/3_satori}

\input{sec/4_vqa_verify}

\input{sec/5_experiments}
\input{sec/6_conclusion}
{
    \small
    \bibliographystyle{ieeenat_fullname}
    \bibliography{main}
}

\input{sec/X_suppl}


\end{document}

%% file: preamble.tex
\usepackage{wrapfig}
\usepackage{algpseudocode}
\usepackage{amsmath}
\usepackage{algorithm}
\usepackage{xcolor}
\usepackage{colortbl}
\usepackage[breakable]{tcolorbox}

\tcbuselibrary{listings}
\tcbuselibrary{breakable}

%% file: sec/0_abstract.tex
\begin{abstract}
DeepSeek-R1 has demonstrated powerful textual reasoning capabilities through reinforcement learning (RL). 
Recent multi-modal studies often directly apply RL to generate R1-like free-form reasoning for multi-modal reasoning tasks.
Unlike textual tasks, multi-modal tasks inherently demand comprehensive visual understanding to effectively address complex challenges.
Therefore, such free-form reasoning faces two critical limitations in these tasks: 
(1) Extended reasoning chains diffuse visual focus away from task-relevant regions, degrading answer accuracy. 
(2) Unverifiable intermediate steps may substantially increase policy-gradient variance and computational costs overhead. 
To this end, we introduce SATORI (\textbf{S}patially \textbf{A}nchored \textbf{T}ask \textbf{O}ptimization with \textbf{R}e\textbf{I}nforcement Learning), which explicitly structures multimodal reasoning process through a Glance-Focus-Think paradigm, converting free-form inference into verifiable reasoning. Specifically, SATORI generates global image captions, and shifts visual attention to task-focus regions vis key bounding boxes, and finally leverages RL over verifiable reasoning patterns to yield the accurate and interpretable answer. 
Furthermore, we introduce VQA‑Verify, a 12k dataset with answer‑aligned captions and bounding boxes to facilitate the three-stage training. 
Experiments demonstrate that SATORI achieves consistent performance improvements across ten multimodal reasoning benchmarks, achieving up to 15.7\% accuracy improvement over R1-like baselines. Our code is available at \href{https://justairr.github.io/SATORI-R1/}{here}.
\end{abstract}

%% file: sec/1_intro.tex
\section{Introduction}
\label{sec:intro}

Nowadays, “Slow-Thinking” multi-modal reasoning models (\eg OpenAI-o1~\cite{openai2024o1}, Gemini~\cite{geminiteam2024gemini15unlockingmultimodal} and DeepSeek-R1~\cite{guo2025deepseek, shao2024deepseekmath}) demonstrate superior performance on complex reasoning tasks (\eg mathematics).
Inspired by DeepSeek-R1~\cite{guo2025deepseek, shao2024deepseekmath}, recent approaches~\cite{kaufmann2023survey, liu2025visual} 
increasingly leverage reinforcement learning (RL) to induce the self-emergence (akin to free-form exploration) of advanced reasoning for complex multimodal tasks~\cite{wang2025multimodalchainofthoughtreasoningcomprehensive, liu2025visual, yu2025dapo}.

\begin{figure*}[th]
  \centering
  \includegraphics[width=\textwidth]{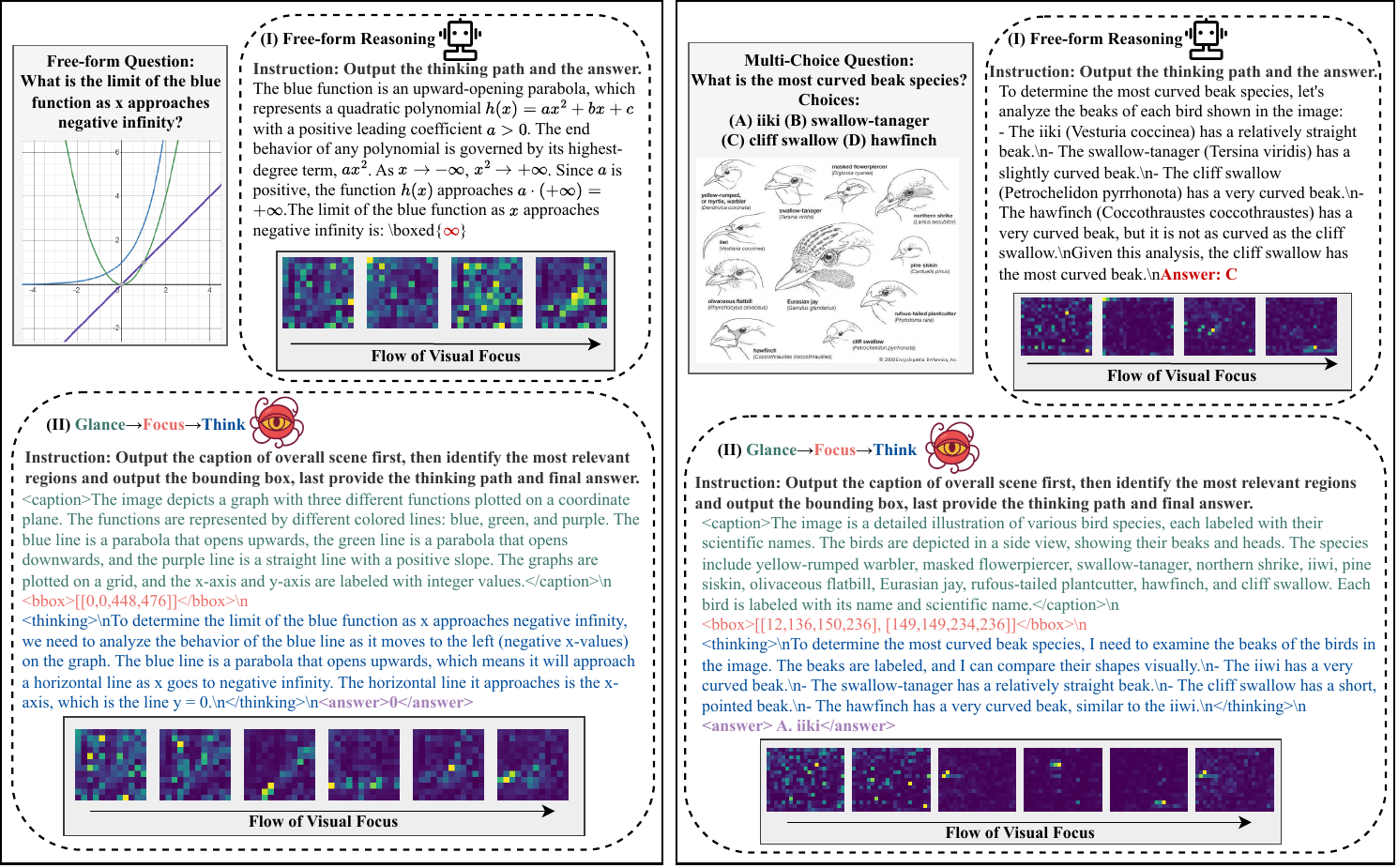}
  \caption{Comparison of Our Reasoning Patterns and Free-form Reasoning. Using the same model Qwen2.5-VL-Instruct-3B with only the output patterns altered, the \textbf{Flow of Visual Focus} heatmaps for free-form reasoning show that attention becomes progressively diffuse and scattered as the reasoning chain lengthens. In contrast, our Glance $\rightarrow$ Focus $\rightarrow$ Think paradigm guides the model's attention from a holistic view to a focused concentration on task-relevant regions. Each attention map is obtained by aggregating approximately 40 tokens output by the model.}
  \label{fig:insight}
\end{figure*}

However, two major limitations hinder applying R1-like reasoning patterns to standard multimodal reasoning tasks: (1) \textbf{Visual-attention Deficiency}: As illustrated in Figure~\ref{fig:insight}, attention analysis reveals that free-form exploration in RL may induce extended reasoning chains that progressively decouple from the image. The visualized attention flow demonstrates that as the text lengthens, the model's focus is diverted from task-relevant regions (such as specific function curves or object details), thus impairing reasoning accuracy;
(2) \textbf{Convergence Impediment}~\cite{havrilla2024teaching, zhang2025grpoleaddifficultyawarereinforcementlearning}: Unstructured reasoning paths not only multiply token consumption but also, in the absence of quantifiable intermediate supervisory verifiable signals, induce high variance in the policy-gradient estimates, thereby slowing convergence. Under standard RL configurations, each training example must undergo multiple rollouts to evaluate an extended reasoning trajectory, further inflating computational overhead.

To bridge these gaps, we propose a structured inference paradigm that strictly enforces visual grounding before logical deduction. Since the model tends to lose focus on task-relevant regions and lacks intermediate supervisory signals, we mandate a sequential workflow: first, scanning the scene for global context (i.e., ``Glance"), then deliberately shifting focus to task-relevant regions (i.e., ``Focus"), and only then synthesizing this focused information to formulate an answer (i.e., ``Think"). Crucially, we utilize these explicit intermediate stages not merely as auxiliary outputs, but as dense, verifiable supervision signals for RL training, ensuring that the reasoning process remains anchored.

To this end, we propose \textbf{SATORI} (\textbf{S}patially \textbf{A}nchored \textbf{T}ask \textbf{O}ptimization with \textbf{R}e\textbf{I}nforcement Learning), a novel RL-optimized structured reasoning paradigm for enhancing MLLM performance. Instead of treating multimodal reasoning as a single black-box mapping from input to answer, SATORI requires the model to follow an explicit \textbf{Glance $\rightarrow$ Focus $\rightarrow$ Think} process. Specifically, SATORI generates global image captions and shifts visual attention to task-relevant regions via key bounding boxes. This explicit spatial grounding fosters sharper attention alignment compared to typical R1-like reasoning patterns. Furthermore, by leveraging these verifiable reasoning patterns (\ie captions and bounding boxes) as intermediate rewards, SATORI provides a smooth approximation for RL optimization, effectively reducing policy-gradient variance by 27\% while yielding accurate and interpretable answers.

In addition, we also introduce VQA-Verify, the first multimodal VQA dataset with both bounding box and caption annotations. It comprises 12k annotated samples across 17 benchmarks, spanning 3 hierarchical categories (\ie Perception, Reasoning, and Multilingual)  and 11 fine-grained task classes, where each including a tuple (image, question, answer), with the corresponding caption describing the image and bounding-box highlighting the answer cue.

We conduct evaluations across seven multimodal reasoning benchmarks, demonstrating that \textbf{SATORI} achieves state-of-the-art performance among models with 7B parameters, improving general visual reasoning benchmarks like MMBench on accuracy by an absolute 8\% over the base model and surpassing comparable methods on mathematical reasoning tasks by 0.9 to 3.3 points.


To summarize, our key contributions are:

\begin{itemize}
  \item We identify a critical failure mode in R1-like multimodal reasoning, termed \textbf{Visual-attention Deficiency}. Through rigorous analysis, we demonstrate that this can be effectively mitigated by our proposed paradigm.
  
  \item We propose a three-step visual reasoning pattern and RL paradigm \textbf{SATORI}. By turning caption and localization into verifiable rewards, our RL paradigm lowers policy-gradient variance by 27\% and speeds up convergence.

  \item We release VQA-Verify, the first augmented dataset of 12k VQA samples with answer-relevant bounding boxes and scene captions to enable explicit supervision.
  
  \item Our method outperforms traditional R1-like free-form reasoning on ten comprehensive benchmarks, achieving up to 15.7\% improvement in accuracy.
\end{itemize}

%% file: sec/2_background.tex
\section{Preliminary}
\subsection{MLLM Architecture and Visual Attention}  
\label{subsec:arch}
Multimodal Large Language Models (MLLMs) unify visual and textual reasoning through a hybrid architecture. Given an input image $\mathbf{I} \in \mathbb{R}^{H \times W \times C}$, a vision encoder (e.g., ViT~\cite{dosovitskiy2020image}) partitions it into $p \times p$ patches, linearly projected into visual tokens $\{\mathbf{v}_i\}_{i=1}^{N_I}$ with $N_I = \frac{HW}{p^2}$. These tokens reside in the same latent space as text tokens $\{\mathbf{t}_j\}_{j=1}^{N_T}$ from language models like Qwen~\cite{bai2023qwen} or Llama~\cite{grattafiori2024llama}.  

The fused sequence $[\mathbf{v}_1, \dots, \mathbf{v}_{N_I}; \mathbf{t}_1, \dots, \mathbf{t}_{N_T}]$ is processed by transformer decoder layers using masked multi-head self-attention (MHSA). For each layer, the attention operation computes:  
\begin{equation}
\operatorname{Attention}(\mathbf{Q}, \mathbf{K}, \mathbf{V}) = \operatorname{softmax}\left(\frac{\mathbf{Q}\mathbf{K}^\top}{\sqrt{d}}\right)\mathbf{V},
\end{equation}  
where $\mathbf{Q}, \mathbf{K}, \mathbf{V}$ are projections of the input sequence. During auto-regressive answer generation, the query $\mathbf{Q}_A$ for each answer token attends to both visual and textual contexts through:  
\begin{equation}
\mathbf{A} = \left[\operatorname{softmax}\left(\frac{\mathbf{Q}_A\mathbf{K}^\top}{\sqrt{d}}\right)\right]_{L,K} \in \mathbb{R}_+^{L \times K \times N_A \times N},
\end{equation}  
where $L$ and $K$ denote the number of layers and attention heads, respectively.

To analyze the visual focus of the models, we isolate the attention weights over visual tokens by reshaping $\mathbf{A}$ into spatial dimensions $(h, w)$, then aggregate multi-head/layer attention of the generated tokens:  
\begin{equation}
\widetilde{\mathbf{A}} = \operatorname{Normalize}\left(\frac{1}{LK}\sum_{l,k}\mathbf{A}_{l,k}\right) \in \mathbb{R}_+^{h \times w}.
\end{equation}  

\subsection{Group Relative Policy Optimization (GRPO)}

Group Relative Policy Optimization (GRPO)~\cite{shao2024deepseekmath} is a reinforcement learning algorithm that optimizes sequence‐generating models without an explicit critic network. For each input \(q\), the current policy \(\pi_{\theta_{\mathrm{old}}}\) samples a group of \(G\) candidate outputs \(\{o_1, \dots, o_G\}\). Each output \(o_i\) receives a reward \(r_i = R(q, o_i)\), and GRPO directly incorporates clipping and KL‐regularization into its objective:

\begin{equation}
\begin{aligned}
\mathcal{J}_{GRPO}(\theta) = & \mathbb{E}[q\sim P(Q),\{o_i\}_{i=1}^G\sim\pi_{\theta_{old}}(O|q)] \\
                          & \frac{1}{G}\sum_{i=1}^G\frac{1}{|o_i|}\sum_{t=1}^{|o_i|} \Biggl\{ \\ 
                          & \quad \min\left[h_{i,t}\hat{A}_{i,t},\operatorname{clip}\left(h_{i,t},1-\varepsilon,1+\varepsilon\right)\hat{A}_{i,t}\right] \\ 
                          & \quad -\beta\mathrm{D}_{KL}\left[\pi_\theta||\pi_{ref}\right] \Biggr\}, 
\end{aligned}
\end{equation}

where $h_{i,t} = \frac{\pi_\theta(o_{i,t}\mid q,\,o_{i,<t})}{\pi_{\theta_{\mathrm{old}}}(o_{i,t}\mid q,\,o_{i,<t})}$ and $\hat{A}_{i,t}$ is the (possibly standardized) advantage at step \(t\). This formulation blends the clipped importance‐sampling term with a KL‐penalty to keep the updated policy close to the reference \(\pi_{\mathrm{ref}}\).

%% file: sec/3_satori.tex
\section{\textit{SATORI}}
\label{sec:satori}
In this section, we first analyze the visual attention maps of the MLLM, demonstrating that different reasoning patterns influence the model’s focus on task-relevant regions and that spatial reasoning patterns enhance attention to key areas (Section~\ref{sec:insight1}). Next, we examine the impact of introducing verifiable reasoning patterns on gradient variance during the RL process (Section~\ref{sec:insight2}). Finally, we propose a visual reinforcement learning paradigm that incorporates verifiable reasoning patterns (Section~\ref{sec:satori_detail}).
\subsection{Spatial Reasoning Patterns Enhance Attention to Task-Relevant Regions}
\label{sec:insight1}



Recent advances~\cite{shao2024deepseekmath, guo2025deepseek} in reinforcement learning, such as in DeepSeek-R1 , have popularized "free-form exploration" reasoning patterns for complex tasks. Inspired by these text-based successes, this paradigm is now being applied to multimodal reasoning tasks.
Although this approach can aid abstract reasoning, multimodal reasoning tasks are intrinsically different and are heavily based on correct understanding of specific image regions. The performance of the model is known to be significantly affected by its attention to these task-relevant regions, as localized attention spikes often correlate with the image areas most relevant to the answer~\cite{zhang2025mllms, wu2024controlmllm}.
However, we find that inference patterns based on free-form reasoning tend to weaken the model's focus on task-relevant regions. This motivates us to explore new forms of reasoning that can guide the model to more accurately attend to key regions of the image, thereby improving the performance of multimodal reasoning.

To quantify the differences in attention distributions under three distinct reasoning paradigms, we randomly sampled 2,000 images from the OpenImages~\cite{kuznetsova2020open} dataset and applied the following inference patterns without any fine-tuning: direct answer, free-form reasoning, and Glance $\rightarrow$ Focus $\rightarrow$ Think. These three represent the inference patterns of the original model, the reasoning-enhanced model, and our proposed method, respectively.
As illustrated in Figure \ref{fig:insight}, we ensured a fair comparison by swapping only the output pipeline and employing a one-shot exemplar to steer the model toward each required format. 
The figure illustrates the flow of the model's visual focus, which is aggregated by averaging over 30-token and 40-token intervals. 

For each generated answer token, we extract the visual attention weights from all layers and heads, and aggregate them into an $h \times w$ grid to obtain the normalized spatial attention distribution $\tilde{A}$. More details could be found at Appendix~\ref{appendix:insight}. Experimental results in Figure~\ref{fig:insight} show that the attention under the Free-Form setting is more dispersed, whereas the Glance $\rightarrow$ Focus $\rightarrow$ Think setting clearly focuses on regions relevant to the question.
Our analysis reveals a critical dependency on \textit{thinking paths}: different reasoning strategies yield distinct attention distributions. As visualized in Figure~\ref{fig:insight}, R1-like free-form reasoning produces scattered attention patterns across decoder layers, with less attention mass concentrated on task-relevant regions during reasoning process. 
This phenomenon may be attributed to the fact that regular multimodal reasoning tasks typically do not require complex chains of reasoning, in contrast to the success of free-form reasoning in more complex mathematical problems. This "overthinking" phenomenon allows the model to hallucinate irrelevant visual features, ultimately diverting focus from semantically salient areas. In contrast, spatial reasoning patterns demonstrate aligning the focus of the layers with human attention.

\begin{figure}[tbh]
  \centering
  \includegraphics[width=0.49\textwidth]{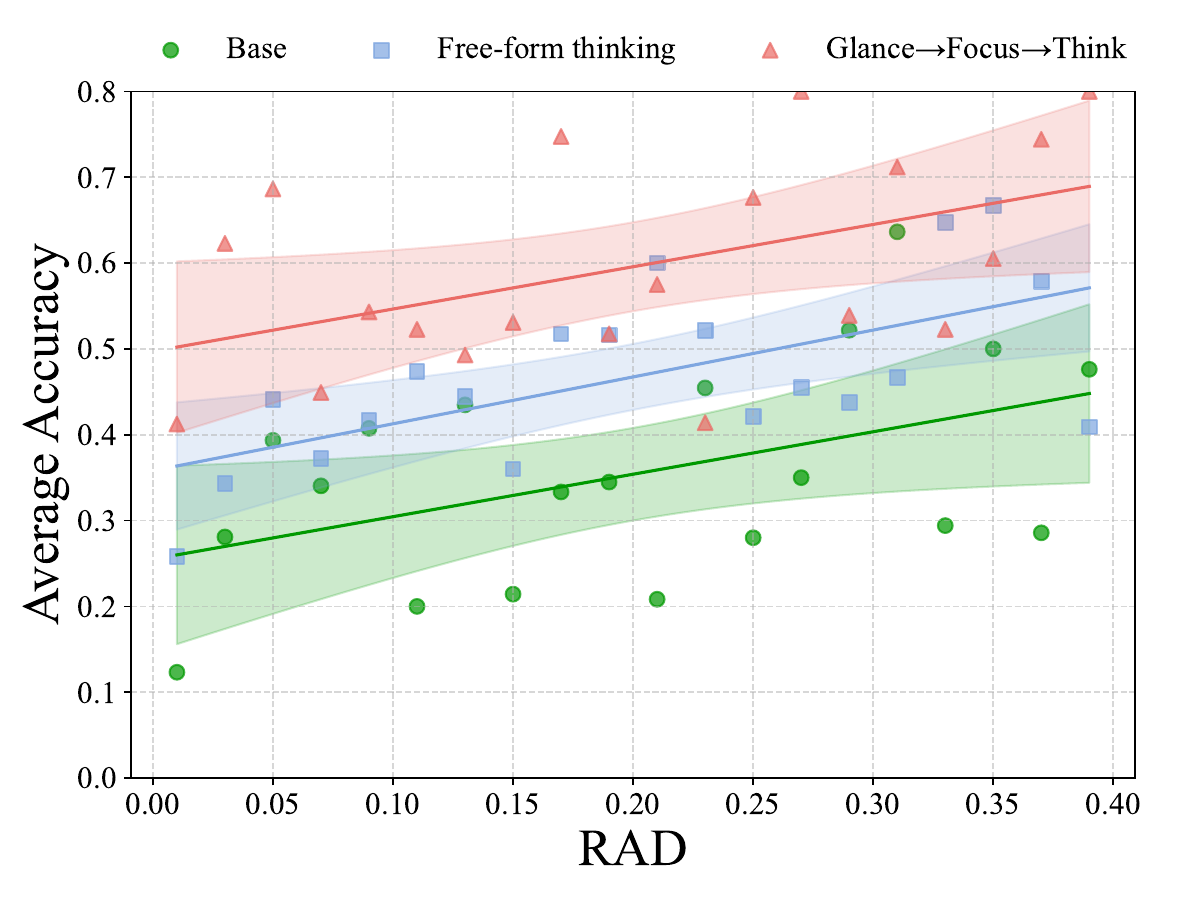}  
  \caption{RAD and accuracy distributions for three different reasoning types. The light‑shaded region represents the 95\% confidence interval.}
  \label{fig:rad_acc}
\end{figure}

This misalignment motivates our quantification framework measuring \textit{Region Attention Density (RAD)}:

\begin{equation}
\text{RAD} = \frac{\sum_{(i,j)\in \mathcal{G}} \widetilde{\mathbf{A}}_{i,j}}{\sum_{i=1}^h \sum_{j=1}^w \widetilde{\mathbf{A}}_{i,j}}
\end{equation} 

where $\mathcal{G}$ is the set of ground truth bounding-boxes. \text{RAD} measures the model’s attention to task-relevant regions by calculating the concentration of the attention map within $\mathcal{G}$. In Figure~\ref{fig:rad_acc}, free-form reasoning patterns exhibit degraded RAD performance due to dispersed attention, whereas our structured \textit{Glance $\rightarrow$ Focus $\rightarrow$ Think} paradigm maintains higher RAD values, with average scores of 0.2621 and 0.2729, respectively. The results also indicate a positive correlation between RAD and accuracy. 
More details can be found in Appendix \ref{appendix:insight}.

Compared to free-form reasoning, our reasoning patterns are also more verifiable and thus better suited as rewards. 

\subsection{Gradient Variance Reduction via Verifiable Reasoning Patterns}
\label{sec:insight2}

\begin{figure*}[ht]
  \centering
  \includegraphics[width=1\textwidth]{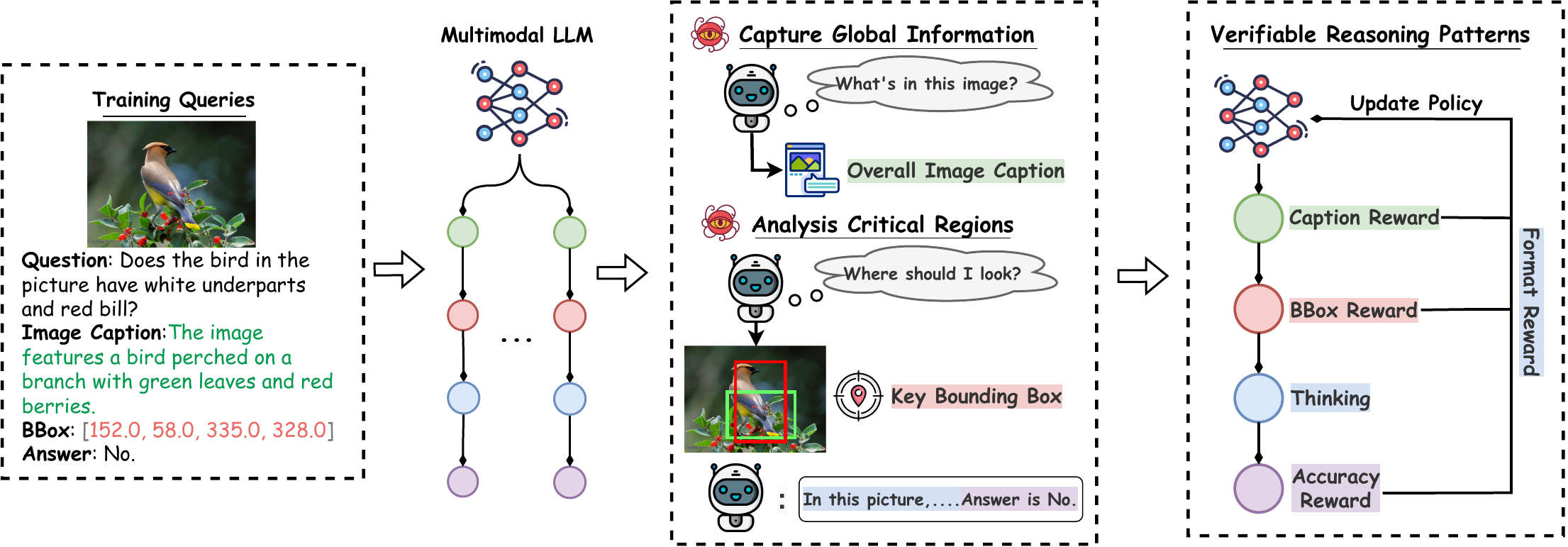}
  \caption{The overview of our proposed method. SATORI guides the model to capture the global information, then analyzes task-relevant regions and finally produces an answer, providing verifiable rewards for step-by-step supervision.}
  \label{fig:overview}
\end{figure*}

Previous studies~\cite{ying2024internlm, shao2024deepseekmath} have demonstrated that combining verifiable reasoning paths with reinforcement learning (RL) yields strong performance on tasks such as mathematical problem solving and logical inference. This success is largely attributed to the availability of well-structured, deterministic reasoning paradigms that allow for step-by-step supervision. In contrast, open-ended multimodal reasoning tasks present significantly higher uncertainty: the reward signals are sparse, the answers are short, and intermediate reasoning steps are typically not explicitly supervised.

These characteristics introduce substantial variance in the estimation of the policy gradient, which poses a major challenge to effective learning~\cite{havrilla2024teaching, zhang2025grpoleaddifficultyawarereinforcementlearning}. In particular, token-level policy gradient methods like GRPO rely on sampled trajectories to estimate gradients, where each trajectory receives a global reward that is distributed uniformly across all tokens. Without verifiable intermediate signals, this global reward is highly variable and often lacks sufficient granularity to guide learning effectively.

Motivated by this issue, we aim to reduce the variance of policy gradients by introducing reasoning patterns that are more stable and verifiable. Instead of relying solely on free-form text reasoning, which is difficult to evaluate and highly stochastic, we design our method to incorporate intermediate reasoning steps that can be evaluated through deterministic criteria. This strategy provides a foundation for smoother gradient estimation, which we analyze in detail in Appendix~\ref{appendix:variance}. We analyze the rationale behind the variance reduction achieved by introducing verifiable rewards in~\ref{appendix:variance}.

\subsection{Spatially Anchored Task Optimization with Reinforcement Learning}
\label{sec:satori_detail}

As stated in Section~\ref{sec:insight1}, we propose a structured and verifiable reasoning pattern that aligns with the intrinsic requirements of multimodal reasoning. 
Specifically, we replace the free-form reasoning with a caption focusing on the overall image and a bounding-box highlighting the key region.
This structured supervision bridges the semantic gap between free-form reasoning and visual grounding requirements. 
\textbf{SATORI} (\textbf{S}patially \textbf{A}nchored \textbf{T}ask \textbf{O}ptimization with \textbf{R}e\textbf{I}nforcement Learning) guides the model to capture both the overall image context and the task-relevant regions before answering the question, providing verifiable rewards for step-by-step supervision.
Figure~\ref{fig:overview} clearly presents the information flow and reward allocation from Caption to BBox to Answer.
VQA-Verify enables direct computation of two verifiable reward signals during RL training:

\begin{align}
\mathcal{R}_{\text{caption}} &= \frac{1}{2} \left( \text{BLEU-4}_{\text{smooth}} + \text{ROUGE-L}_{\text{F1}} \right), \label{eq:caption} \\
\mathcal{R}_{\text{bbox}} &= \text{Union IoU}(\mathcal{P}, \mathcal{G}) \label{eq:bbox}
\end{align}

where $\mathcal{P}$ is the set of predicted boxes and $\mathcal{G}$ is the set of ground-truth bound boxes. The caption reward $\mathcal{R}_{\text{caption}}$ combines smoothed BLEU-4 with ROUGE-L F1.
Since there may be multiple bounding-boxes in the ground-truth, we define the \textbf{Union IoU} to compute the intersection over union of the combined bounding-boxes.
The detailed calculation process can be found in Algorithm~\ref{alg:union_iou}.
Similar to the R1-like reasoning training method, we also sample two types of reward signals during training: the accuracy reward $R_{\text{acc}}$ and the format reward $R_{\text{format}}$. The accuracy reward $R_{\text{acc}}$ measures whether the generated answer matches the ground-truth, while the format reward $R_{\text{format}}$ ensures that the output follows the expected format of Glance $\rightarrow$ Focus $\rightarrow$ Think. Both reward signals take binary values of 0 or 1.

\begin{algorithm}[thb]
\caption{Union IoU Reward Computation}
\label{alg:union_iou}
\begin{algorithmic}
\Require
Predicted boxes: $\mathcal{P} = \{B_i^p\}_{i=1}^{N_p}$ where $B_i^p = [x_1^i, y_1^i, x_2^i, y_2^i]$ \\
Ground-truth boxes: $\mathcal{G} = \{B_j^g\}_{j=1}^{N_g}$ where $B_j^g = [x_1^j, y_1^j, x_2^j, y_2^j]$

\Ensure IoU score: $\mathcal{R}_{\text{bbox}} \in [0,1]$

\State Convert boxes to geometric polygons: 

$\begin{cases}
\mathcal{P}_{poly} = \bigcup_{i=1}^{N_p} \text{Rect}(x_1^i, y_1^i, x_2^i, y_2^i) \\ 
\mathcal{G}_{poly} = \bigcup_{j=1}^{N_g} \text{Rect}(x_1^j, y_1^j, x_2^j, y_2^j)
\end{cases} 
$

\Comment{$\text{Rect}(a,b,c,d)$: Axis-aligned rectangle defined by coordinates $(a,b)$ and $(c,d)$.}

\State Compute union regions:
$
\begin{cases}
\mathcal{U}_p = \text{Union}(\mathcal{P}_{poly}) \\
\mathcal{U}_g = \text{Union}(\mathcal{G}_{poly})
\end{cases}
$

\State Calculate intersection and union areas:
$
\begin{cases}
A_{\cap} = \text{Area}(\mathcal{U}_p \cap \mathcal{U}_g) \\
A_{\cup} = \text{Area}(\mathcal{U}_p \cup \mathcal{U}_g)
\end{cases}
$

\State Compute final IoU with numerical stability:
$
\mathcal{R}_{\text{bbox}} = \frac{A_{\cap}}{A_{\cup} + \epsilon} \quad (\epsilon = 10^{-6})
$
\end{algorithmic}
\end{algorithm}





Subsequently, we guide the model within the prompt to first reason about the image caption and the key region bounding-box, and optimize the model solely using the GRPO paradigm. Compared to SFT, which simply imitates annotated data, this training approach leverages policy gradients to encourage the model to explore better generation strategies. By employing independent reward functions for each subtask, the model receives explicit feedback to optimize final accuracy. Similar to the setup in GRPO, we also adopt the basic format reward and accuracy reward.

%% file: sec/4_vqa_verify.tex
\section{{VQA-Verify Dataset}}
\label{sec:vqa-verify}

To address the scarcity of explicit visual supervision in multimodal reasoning training, we introduce VQA-Verify, an augmented dataset providing verifiable grounding signals for 12,000 samples across standard multimodal reasoning benchmarks. Different from standard VQA datasets, VQA-Verify provides annotations for each sample in the form of $\left(\text{Image}, \text{Question}, \text{Caption}, \text{BBox}, \text{Answer}\right)$. To the best of our knowledge, VQA-Verify is the first multimodal dataset that annotates bounding-box and image caption.

\begin{figure}[th]
  \centering
  \includegraphics[height=220pt]{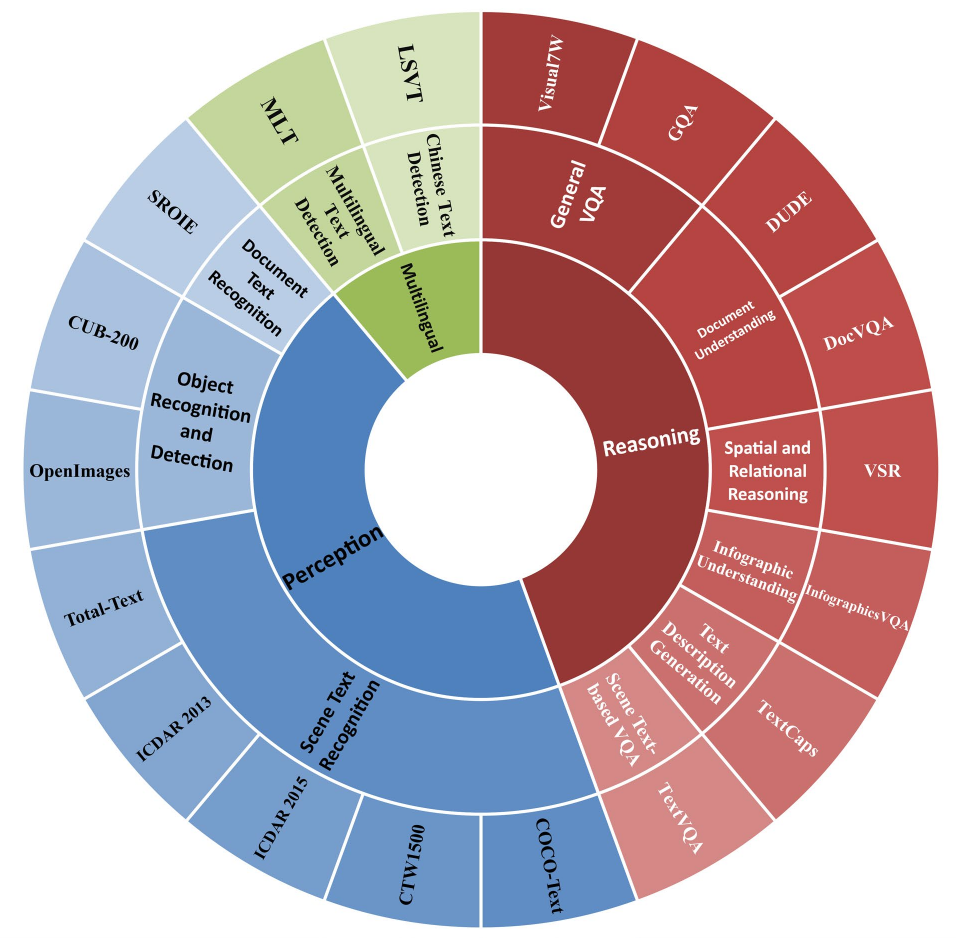}
  \caption{Overview of VQA-Verify. VQA-Verify is divided into 3 categories, 11 subtasks, and 17 benchmarks in total.}
  \label{fig:VQA-Verify}
\end{figure}

Inspired by previous works~\cite{liu2024mmbench, chen2024we}, the novel VQA-Verify framework integrates an extensive collection of 17 benchmark datasets through a hierarchical framework that is specifically designed to address diverse visual-textual understanding capabilities. At the highest level, the dataset spans three primary categories: \textbf{Perception}, \textbf{Reasoning}, and \textbf{Multilingual} tasks.

\textbf{Perception.} This category focuses on foundational visual-textual recognition through two subcategories. Document Text Recognition is supported by SROIE~\cite{huang2019icdar2019}, which specializes in scanned receipt OCR and key information extraction. Scene Text Recognition encompasses six datasets: Total-Text~\cite{ch2020total} for multi-oriented and curved text, ICDAR 2013~\cite{karatzas2013icdar} and ICDAR 2015~\cite{karatzas2015icdar} as standard benchmarks for horizontal and scene text detection, CTW1500~\cite{liu2019curved} for curved text analysis, and COCO-Text~\cite{veit2016coco} for text detection in complex scenes. Object Recognition and Detection integrates CUB-200~\cite{wah2011caltech} for fine-grained bird classification and OpenImages~\cite{kuznetsova2020open} for large-scale object detection with bounding-boxes and visual relationships.

\textbf{Reasoning.} This category targets advanced cognitive tasks through six subdomains. Scene Text-based VQA leverages TextVQA~\cite{singh2019towards} for question answering requiring textual reasoning in images, while Text Description Generation uses TextCaps~\cite{sidorov2020textcaps} to challenge models in generating context-aware captions that fuse text and visual elements. Document Understanding combines DocVQA~\cite{mathew2021docvqa} and DUDE~\cite{van2023document} to evaluate multi-page document comprehension and layout-aware reasoning. Infographic Understanding employs InfographicVQA~\cite{mathew2022infographicvqa} to test joint analysis of graphical layouts, data visualizations, and textual content. General VQA integrates GQA~\cite{hudson2019gqa} for balanced question-answering with scene graph support and Visual7W~\cite{zhu2016visual7w} for object-grounded multimodal QA. Lastly, Spatial and Relational Reasoning incorporates VSR~\cite{liu2023visual} to assess spatial relation verification between objects.

\textbf{Multilingual.} This category includes LSVT~\cite{sun2019chinese} for Chinese text detection in street-view scenarios and MLT~\cite{nayef2019icdar2019} for script-agnostic text detection across diverse languages. This hierarchical integration provides a framework for evaluating both fundamental perception skills and sophisticated reasoning abilities across monolingual and cross-lingual contexts, while maintaining alignment with real-world challenges through its constituent datasets.

The bounding-boxes in the dataset are derived from the works of \citet{shao2024visual} and \citet{wang2020general}. We employed GPT-4o, one of the state-of-the-art models, for image captioning. Following this process, we conducted manual quality review and refinement, and further integrated the VQA-Verify dataset. A more detailed description of the datasets is provided in Appendix \ref{appendix:dataset}.

%% file: sec/5_experiments.tex
\begin{table*}[tb]
  \centering
  \caption{Comparison with other reasoning models on eight multimodal reasoning datasets. The results indicate that our method maintains competitive performance on diverse multimodal reasoning benchmarks.}
  \label{tab:math-benchmarks}
  \setlength\tabcolsep{2pt}
  \renewcommand{\arraystretch}{1.1}
  \scalebox{1.0}{
    \begin{tabular}{lcccccccc}
      \toprule
      Method                      & MathVista & Math‑V & MathVerse & OlypamidBench & WeMath & MMStar & MMBench & MMMU  \\ 
      \hline
      \rowcolor{gray!10}
      \multicolumn{3}{l}{\textit{Closed-Source Model}} & & & & & &\\
      GPT-4o~\cite{hurst2024gpt} & 63.8 & 30.3 & 39.4  & 35.0 & 68.8 &  65.1 & 84.3 &	70.7 \\
      Claude-3.5 Sonnet~\cite{claude_3.5_sonnet} & 61.8 & 38.0 & -  & - & - & 65.1 & 81.7 & 66.4 \\
       \hline
      \rowcolor{gray!10}
      \multicolumn{3}{l}{\textit{Open-Source General Model (2-3B)}} & & & & & & \\
      Qwen2.5-VL-3B~\cite{bai2025qwen2} & 61.2 & 21.2 & 47.6 & 10.3 & 22.1 &  56.3 & 60.8 & 51.2 \\
      InternVL3-2B~\cite{zhu2025internvl3} & 57.6 & 21.7  & 25.3  & 9.6 & 22.4 & \textbf{61.1} & \textbf{78.0} & 48.7 \\
        \hline
      \rowcolor{gray!10}
      \multicolumn{3}{l}{\textit{Open-Source Reasoning Model (2-3B)}} & & & & & & \\
      R1‑VL‑2B~\cite{zhang2025r1}           & 52.1      & 17.1   & 26.2 & - & - & 49.8 & - & -   \\
      Aquila-VL-2B~\cite{gu2025infinitymmscalingmultimodalperformance}     & 59.0 & 18.4 & 26.2 & - & - & 54.9 & 75.2 & 46.9  \\
      InternVL2.5-2B-MPO~\cite{chen2024expanding} & 53.4 & - & - & - & - & 54.9 & 70.7 & 44.6 \\
      VLAA-Thinker-3B~\cite{chen2025sftrlearlyinvestigation} & 61.0 & 24.4 & 36.4 & - & 23.2 & - & - & - \\
        \hline
      \rowcolor{gray!10}
      \multicolumn{3}{l}{\textit{Our Model (3B)}} & &  & & & & \\
        \textbf{SATORI-3B w/o thinking}                     & 60.9      & 21.7   & 32.2  &   10.9 & 25.6 &  55.9 &  76.5 & 54.7   \\
        \textbf{SATORI-3B}                     & \textbf{67.4}      & \textbf{26.1}   & \textbf{39.8}  &   \textbf{13.5} & \textbf{30.1} &  \textbf{56.7} &  76.9 & \textbf{56.9}   \\
       \hline
      \rowcolor{gray!10}
      \multicolumn{3}{l}{\textit{Open-Source General Model (7-11B)}} & & & & & & \\
      InternVL2.5-8B~\cite{chen2024expanding} & 64.4 & 19.7 & 39.5 & 12.3 & 53.5 & 63.2 & 82.5 & 56.2 \\
      InternVL3-8B~\cite{zhu2025internvl3} & 71.6 & 29.3 & 39.8 & - & 37.1 & 68.7 & 82.1 & 62.2 \\
      Qwen2.5-VL-7B~\cite{bai2025qwen2} & 68.2 & 25.4 & 47.9 & 20.2 & 62.1 & 64.1 & 82.2 & 58.0 \\
      \hline
      \rowcolor{gray!10}
      \multicolumn{3}{l}{\textit{Open-Source Reasoning Model (7-11B)}}& & & & & & \\
      Adora-7B~\cite{anonymous2025adora} & 73.5 & 23.0 & 50.1 & 20.1 & 64.2 & - & - & - \\
      InternVL2.5-8B-MPO~\cite{chen2024expanding} & 68.9 & 21.5 & 35.5 &  7.8 &  53.5 & 62.5 & 76.5 & - \\
      R1-Onevision-7B~\cite{yang2025r1} &  64.1 & 23.5 & 47.1 & 17.3 &  61.8 & - & - & - \\
      OpenVLThinker-7B~\cite{deng2025openvlthinker} & 70.2 & 25.3 & 47.9 & 20.1 & 64.3 & - & - & - \\
      MM-Eureka-7B~\cite{meng2025mm} & 73.0 & 26.9 & 50.3 & 20.1 & \textbf{66.1} & - & - & - \\
      VL-Rethinker-7B~\cite{wang2025vlrethinkerincentivizingselfreflectionvisionlanguage} & 73.7 & 30.1 & 54.6 & - & - & - & - & 56.7 \\
      MMR1-7B~\cite{leng2025mmr1enhancingmultimodalreasoning} & 72.0 & 31.8 & 55.4 & - & - & - & - & - \\
      \hline
      \rowcolor{gray!10}
      \multicolumn{3}{l}{\textit{Our Model (7B)}} & & & & & & \\

          \textbf{SATORI-7B w/o thinking} & 71.3 & 30.2 & 49.2 & 20.4 & 64.1 & \textbf{69.7} & 82.0 & 60.6 \\
      \textbf{SATORI-7B} & \textbf{76.2} & \textbf{32.7} & \textbf{56.9} & \textbf{23.7} & 65.2 & 69.5 & \textbf{82.9} & \textbf{63.6} \\
      \bottomrule
    \end{tabular}
  }
\end{table*}

\section{Experiments}
\subsection{Implementation}
\label{sec:implementation}
Our model is based on the Qwen2.5-VL-Instruct-3B and Qwen2.5-VL-Instruct-7B backbone. We perform direct RL training using the framework introduced in ~\cite{zhao2025swift} without any cold start. 
The training approach adopts the GRPO-zero~\cite{shao2024deepseekmath} method, with the reward function aligned with that of Section~\ref{sec:satori}. For training data, we use the lightweight VQA-Verify dataset proposed earlier in Section~\ref{sec:vqa-verify}. The model uses a configuration of $256\times28\times28$ as the min pixel setting and $512\times28\times28$ as the max pixel setting. 
More implementation details could be found in Appendix~\ref{appendix:implementation}. 

For evaluation, we primarily rely on several comprehensive benchmarks: MMBench~\cite{liu2024mmbench}, MMStar~\cite{chen2024we}, MMMU~\cite{yue2024mmmumassivemultidisciplinemultimodal}, MME~\cite{fu2024mmecomprehensiveevaluationbenchmark} and OCRBench~\cite{liu2024ocrbench}. We also compare our method with the current state-of-the-art reasoning models on five mathematical datasets: MathVista~\cite{lu2023mathvista}, Math-V~\cite{wang2024measuring}, MathVerse~\cite{zhang2024mathverse}, OlypamidBench~\cite{he2024olympiadbench} and WeMath~\cite{qiao2025we}. Results are presented in Table~\ref{tab:math-benchmarks} and Figure~\ref{fig:mme}.

\subsection{Main Results}
As detailed in Table~\ref{tab:math-benchmarks}, our larger SATORI-7B model further widens the performance gap, establishing new state-of-the-art results among open-source models across numerous reasoning benchmarks. Most notably, on the comprehensive MathVista benchmark, SATORI-7B achieves an outstanding 76.2\%. This result not only significantly outperforms all other open-source reasoning models, including the next-best Adora-7B (73.5\%), but also surpasses leading closed-source systems like GPT-4o (63.8\%) and Claude-3.5 Sonnet (61.8\%). This superior performance is consistent across other challenging datasets: SATORI-7B achieves the top open-source scores on MMMU (63.6\%), MathVerse (50.9\%), MMBench (82.9\%), and OlypamidBench (20.7\%). Furthermore, on the highly difficult Math-V dataset, SATORI-7B (32.7\%) robustly outperforms all open-source competitors and significantly narrows the gap to the top closed-source model (GPT-4o at 30.3\%). These results collectively validate the scalability and exceptional reasoning capabilities of our SATORI framework.

\begin{figure}[tb]
  \centering
  \includegraphics[width=\linewidth]{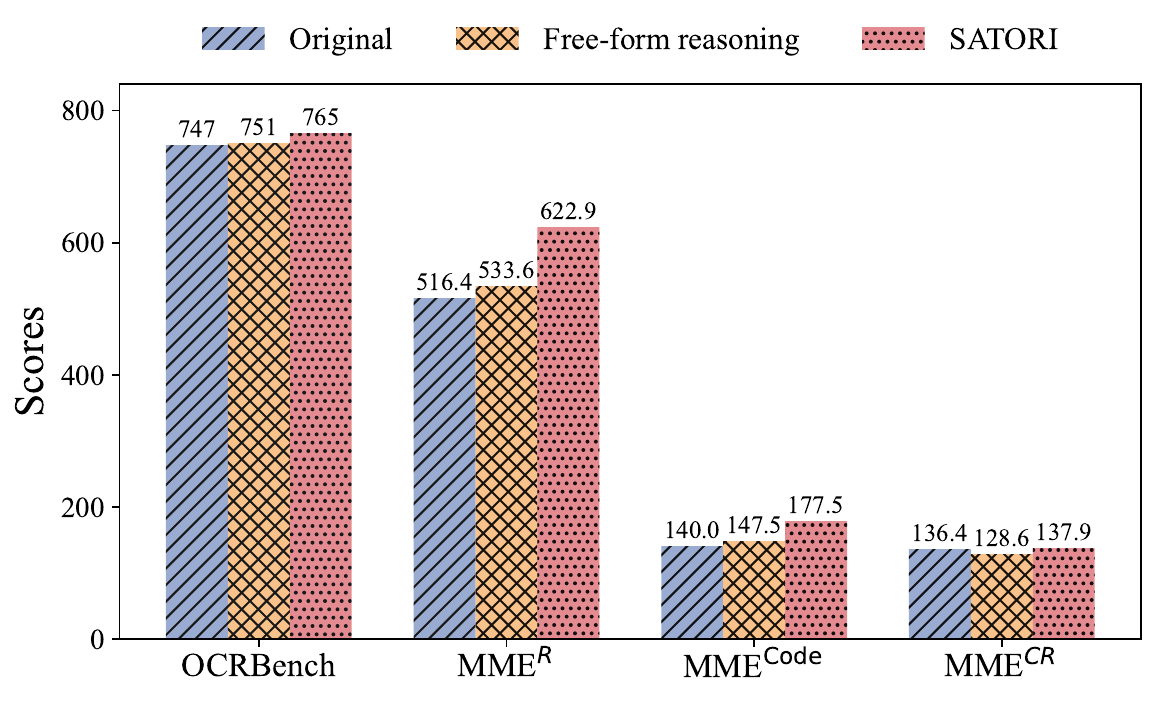}
  \caption{Performance of different methods across various reasoning and OCR benchmarks on Qwen-2.5-VL-Instruct-3B. Specifically, MME$^{R}$, MME$^{\text{Code}}$ and MME$^{CR}$ denote the Reasoning, Code Reasoning and Commonsense Reasoning of MME~\cite{fu2024mmecomprehensiveevaluationbenchmark}.}
  \label{fig:mme}
\end{figure}

As illustrated in Figure~\ref{fig:mme}, SATORI consistently outperforms both the original Qwen2.5-VL-Instruct-3B baseline and its free-form reasoning variant across all tasks. The performance gains are especially significant on MME Reasoning (MME$^{R}$), where SATORI (622.9) substantially surpasses the original model (516.4), and on MME Code Reasoning (MME$^{Code}$), where SATORI achieves 177.5 compared to the baseline's 140.0. Additional results and analyses can be found in Appendix~\ref{appendix:more_experiments}.

\subsection{Additional Experiments}

\textbf{Ablation Study on Reasoning Patterns.}
To validate the effectiveness of each component in SATORI, we compare the model's performance under different reward configurations: using BBox without Caption, using neither, and supervised fine-tuning. All experiments were conducted on the same VQA-Verify dataset and Qwen-2.5-VL-Instruct-3B, with hyperparameter settings consistent with those described in Appendix~\ref{appendix:implementation}. The only differences are in the training strategies and the choice of reward signals.

The results in Table~\ref{tab:ablation} demonstrate that our method achieves the best performance when both BBox and Caption reward signals and thinking section are present. 

\begin{table}[th]
  \centering
  \caption{Ablation results on reasoning patterns.}
  \label{tab:ablation}
    \small
  \begin{tabular}{lcc}
    \toprule
    \textbf{Method} & \textbf{MMBench} & \textbf{MMStar} \\
    \midrule
    \rowcolor{gray!10}
      Qwen2.5-VL-Ins-3B            & 60.8 & 48.0 \\
      +BBox+SFT                    & 58.9 & 49.7 \\
      +BBox+Caption+SFT            & 65.2 & 50.5 \\
      +Free Form Reasoning+RL      & 64.6 & 50.4 \\
      +BBox+RL                     & 71.0 & 54.1 \\
      +BBox+Think+RL & 73.3 & 54.4 \\
      +Caption+RL & 63.0	& 51.5 \\
      +Caption+Think+RL & 63.8 & 53.5 \\
      +BBox+Caption+RL & 76.5 & 55.9 \\
    \midrule
      SATORI                       & \textbf{76.9} & \textbf{56.1} \\
    \bottomrule
  \end{tabular}
\end{table}

\begin{table}[tb]
\centering
\small
\caption{Performance comparison on the InternVL3-2B model, where FFR denotes Free-form Reasoning.}
\label{tab:internvl3-comparison}
\begin{tabular}{lcccc}
\toprule
Method & MMBench & MMStar & MMMU & MathVista \\
\midrule
Original & 78.6 & 61.1 & 48.7 & 57.5 \\
FFR & 79.0 & 62.6 & 50.2 & 57.2 \\
\textbf{SATORI} & \textbf{80.7} & \textbf{65.9} & \textbf{52.8} & \textbf{59.0} \\
\bottomrule
\end{tabular}
\end{table}

\textbf{Experiments on More Model Families.}
To verify the model's generalization ability, we implemented our method on one of the Vision LLMs, InternVL3-2B. The results in Table~\ref{tab:internvl3-comparison} indicate that our method generalizes well across different model families.

\textbf{Variance Reduction Analysis.}
We compared the policy variance over training epochs between our approach and the free-form reasoning baseline. This experiment was conducted on Qwen-2.5-VL-Instruct-3B, maintaining the same training dataset and parameters.

\begin{figure}[htbp]
    \centering
    \begin{subfigure}[b]{0.49\textwidth}
  \centering
  \includegraphics[width=\textwidth]{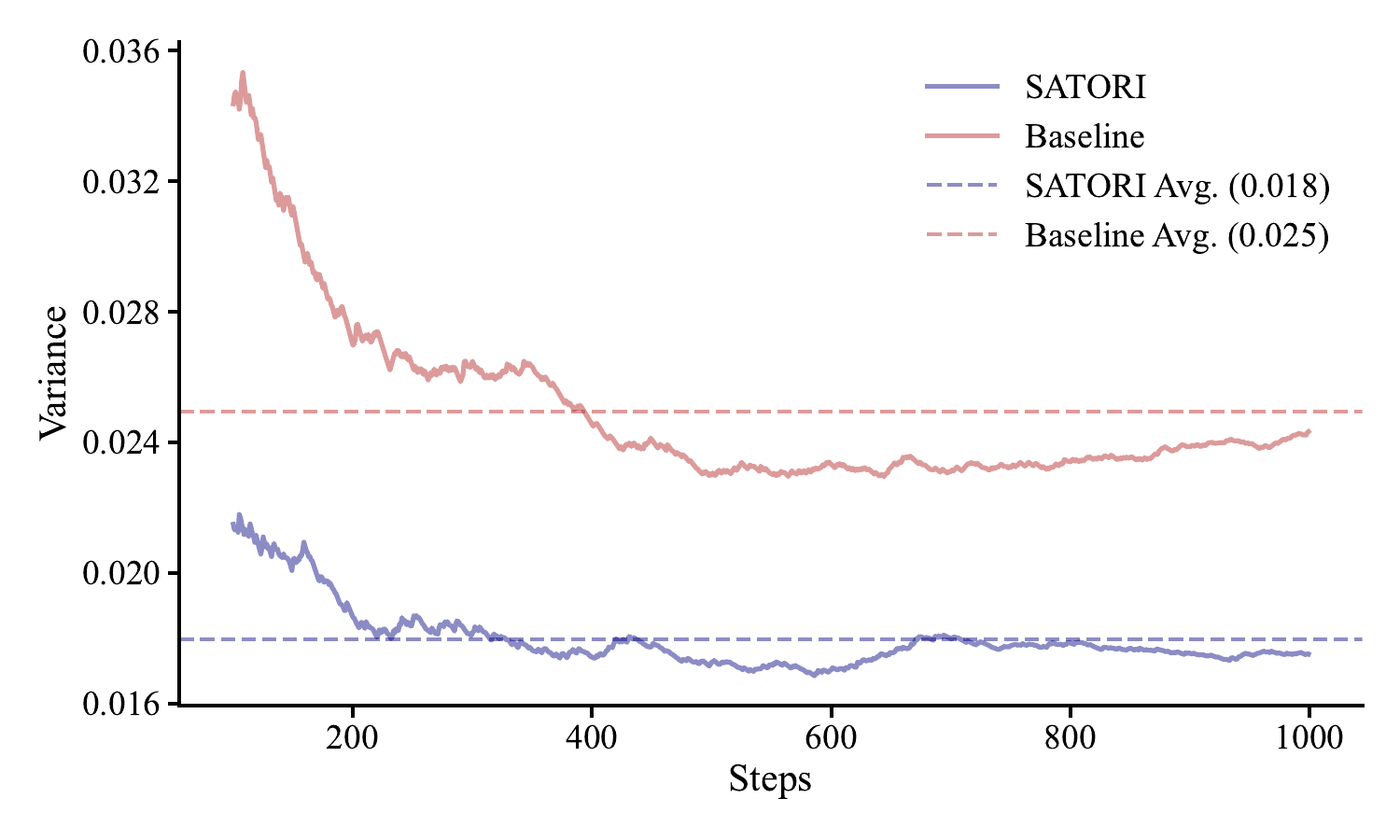} 
  \caption{}
  \label{fig:varaiance}
    \end{subfigure}
    \hfill
    \begin{subfigure}[b]{0.49\textwidth}
        \centering
        \includegraphics[width=\textwidth]{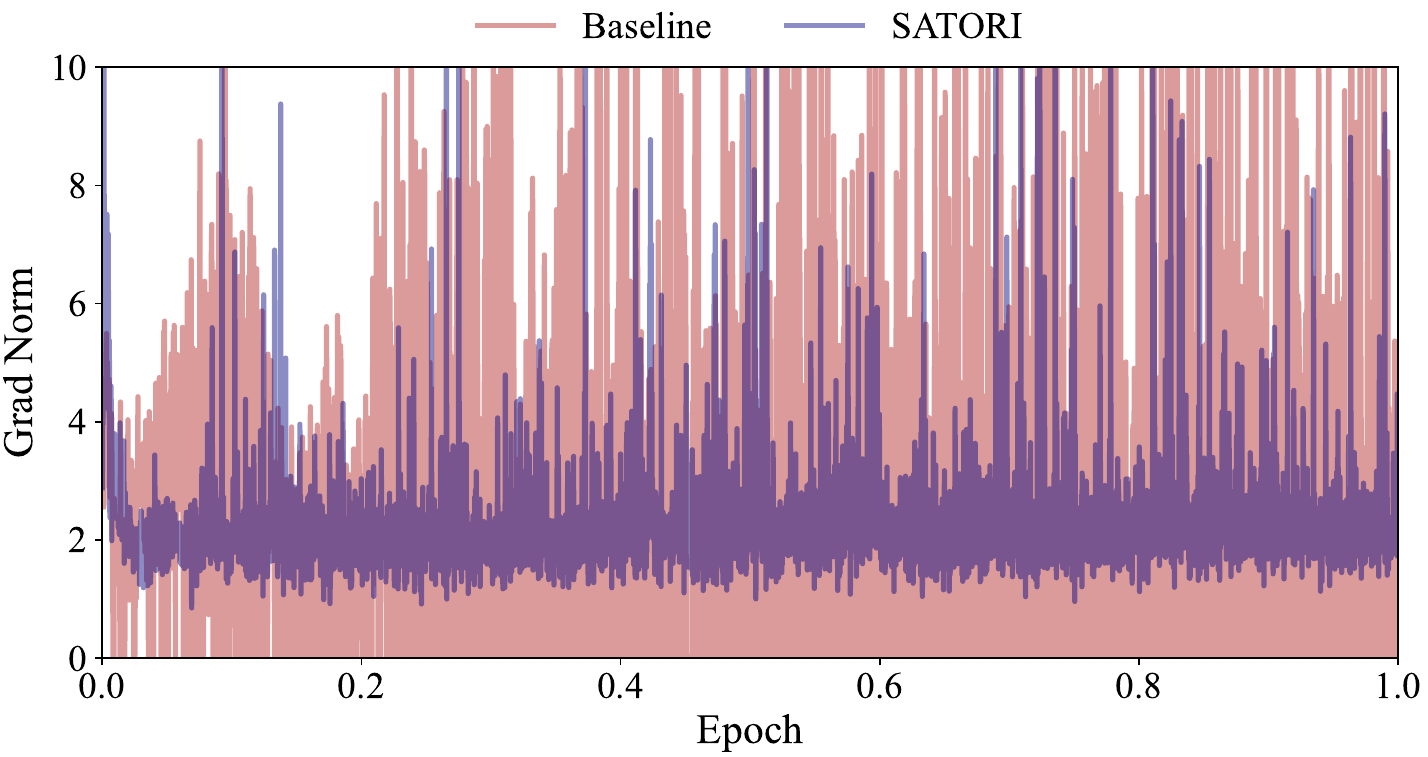}
        \caption{}
        \label{fig:grad_norm}
    \end{subfigure}
    \caption{\textbf{a)} Comparison of SATORI with the classic free-form reasoning approach reveals the variance in GRPO performance. \textbf{b)} Changes in gradient norm over training epochs. The results show that our method converges faster than free-form reasoning.}
    \label{fig:combined}
\end{figure}

As shown in Figure~\ref{fig:varaiance}, SATORI exhibits substantially lower gradient variance compared to free-form reasoning baselines. The average variance during training drops from 0.025 to 0.018. This suggests that verifiable intermediate rewards act as variance-reducing control signals, enabling more stable and efficient policy learning. Moreover, the gradient‐norm curves in Figure~\ref{fig:grad_norm} show that SATORI converges in fewer steps, confirming that variance reduction translates directly into faster training.

%% file: sec/6_conclusion.tex
\section{Conclusion}

In this paper, we presented \textbf{SATORI}, a structured reasoning paradigm designed to enforce spatial grounding through an explicit \textbf{Glance $\rightarrow$ Focus $\rightarrow$ Think} process. It addresses the attention dilution and high gradient variance of free-form reasoning by decomposing tasks into three verifiable steps: caption generation, region localization, and answer prediction. This approach uses intermediate rewards to sharpen visual focus and reduce policy-gradient variance by 27\%. To support this method, we introduced the VQA-Verify dataset, which provides the necessary explicit supervision. Extensive experiments show SATORI achieves significant performance gains, including a 15.7\% absolute improvement on MMBench.
Overall, this work demonstrates that by optimizing an explicit, spatially anchored reasoning process, we can build more effective multimodal models.

%% file: sec/X_suppl.tex
\clearpage
\setcounter{page}{1}
\maketitlesupplementary

\section{Related Works}
\label{appendix:related_works}

\subsection{Enhancing Reasoning in MLLMs}
Multimodal Large Language Models (MLLMs) have rapidly advanced the state of vision–language understanding by integrating text, image, and other sensory inputs. Early efforts primarily tackled image–text tasks, demonstrating the ability to generate descriptive captions and answer visual queries based on single-image prompts \cite{llava,llavanext}. Subsequent research extended these models to video understanding \cite{sun2024video,cheng2024videollama} and to more diverse modalities such as audio and point clouds \cite{lyu2023macaw,wu2023next}. Domain-specific adaptations have further pushed the envelope: examples include specialized medical image interpretation \cite{zhang2023pmc,li2023llavamed,lan2025gem} and structured document analysis \cite{ye2023mplugdoc,liu2024textmonkey}. 

Building on the success of chain-of-thought prompting in pure-language settings \cite{openai2024o1}, recent approaches seek to endow MLLMs with stronger inference capabilities via supervised fine-tuning on high-quality reasoning traces. Methods such as LLaVA‑CoT generate structured intermediate reasoning steps—summary, description, analysis, conclusion—using a powerful teacher model and then train the target MLLM on these examples \cite{xu2024llava}. Other works incorporate search techniques: Mulberry employs a collective Monte Carlo Tree Search across multiple model instances to discover effective reasoning pathways, which are subsequently distilled into a single model \cite{yao2024mulberry}. However, unlike these methods that rely on imitation, SATORI introduces verifiable visual grounding signals within a reinforcement learning framework to explicitly align reasoning with image content, rather than the free-from thinking applied in normal RL.

\subsection{Reinforcement Learning for Structured Reasoning}
Reinforcement Learning (RL) provides a principled approach for sequential decision-making, where agents optimize long-term return through trial-and-error interactions \cite{kaelbling1996reinforcement,watkins1992q}. In the context of large language models, RL with human feedback (RLHF) has been instrumental in aligning generation quality to human preferences, using algorithms like PPO \cite{PPO} and DPO \cite{DPO} \cite{bai2022training}. More recently, RL has been adopted to improve reasoning: ReST‑MCTS* introduces a learned process reward model to evaluate intermediate reasoning steps \cite{zhang2024rest}, while other studies demonstrate that simple outcome-level rewards—assigning positive credit only to sequences that reach correct answers—are sufficient to guide policy optimization \cite{luong2024reft,guo2025deepseek,team2025kimi,huang2025visionr1incentivizingreasoningcapability}. In contrast to these free-form reasoning paradigms which may lead to attention dilution in multimodal contexts, SATORI employs a structured Glance-Focus-Think process with intermediate verifiable rewards to maintain visual focus and reduce gradient variance.

\section{Details of Visual Attention Map Comparison}
\label{appendix:insight}
\subsection{Implementation}
To ensure a fair comparison, we employed the original Qwen2.5-VL-Instruct-3B model to analyze visual attention maps under three different reasoning patterns, without any additional fine-tuning. The experiments were conducted on 2,000 randomly selected samples from the OpenImages~\cite{kuznetsova2020open} dataset.

Since the instruction-following capability of the 3B model is relatively limited, simply prompting it with a reasoning pattern may not guarantee adherence. Therefore, we adopted a one-shot example setting for comparative experiments. The prompts used in the experiments are as follows:

\begin{tcolorbox}[colback=gray!5!white,colframe=black!75!black,title=Prompt for Visual Attention Map Comparison,breakable]
\begin{verbatim}
"ConventionalVQA": 
# Output Example
Question: What is the capital city 
of France?
Answer: Paris
------------------
"Free_Form_Reasoning":
Output the thinking process in 
<think> and final answer in 
<answer> </answer> tags.

# Output Example
Question: What is the capital city 
of France?
<think>France is a country in 
Europe, and its capital city is
Paris. </think>
<answer>Paris</answer>
------------------
"Caption_BBox_Answer": 
First, provide an image caption 
describing the overall scene inside 
<caption> </caption>. Then, output
the list of bounding-boxes in the 
format of [[x1,y1,x2,y2], ...]
inside <bbox> </bbox>. Finally, give
the final answer in <answer>
</answer>.

# Output Example
Question: What is shown in the
image?
<caption>A group of people playing
soccer on a green field.</caption>
<bbox>[[50,60,120,180], [200,80,
260,190], [300,90,360,200]]</bbox>
<answer>People are playing soccer.
</answer>
------------------
"SATORI":
First, generate a brief image
caption describing the overall
scene. Provide the caption inside
<caption> </caption>.

Next, identify the most relevant
image regions for answering the
question. Enclose these coordinates
in <bbox>[[x1,y1,x2,y2], ...]
</bbox>.

Then, formulate a step-by-step
thinking process that outlines
the reasoning required to arrive
at the solution. Enclose this
reasoning in <thinking>
</thinking> tags.

Finally, provide the final answer to
the question inside <answer>
</answer> tags.

\end{verbatim}
\end{tcolorbox}

\subsection{Visual Attention Map Computation in MLLMs}
Similar to the work of~\citeauthor{zhang2025mllms}, when the model generates the $n$-th answer token during autoregressive decoding, we first extract the attention tensor across all layers and heads:

\begin{equation}
  \mathbf{A} = \bigl[\mathrm{softmax}\!\bigl(\tfrac{\mathbf{Q}_A \mathbf{K}^\top}{\sqrt{d}}\bigr)\bigr]_{L\times K} \in \mathbb{R}_+^{L\times K\times N_A\times N},
\end{equation}

where $L$ is the number of layers, $K$ is the number of heads per layer, $N_A$ is the number of answer‐token queries, and $N$ is the total length of text and visual tokens. We then locate the start and end indices of the visual tokens in the input sequence, denoted $\mathrm{vs\_pos}$ and $\mathrm{ve\_pos}$, and crop each layer–head attention to the visual embedding segment:
\begin{equation}
  \mathbf{A}_I^{(\ell,k)} = \mathbf{A}^{(\ell,k)}_{n,\;\mathrm{vs\_pos}:\mathrm{ve\_pos}} \in \mathbb{R}_+^{N_A\times HW},
\end{equation}
where $HW$ is the flattened spatial length of the visual patches. This segment is reshaped into a two‐dimensional grid:
\begin{equation}
  \mathbf{A}_I^{(\ell,k)} \xrightarrow{\;\mathrm{reshape}\;} \tilde{\mathbf{A}}^{(\ell,k)} \in \mathbb{R}_+^{h\times w},
\end{equation}
with $h$ and $w$ denoting the number of patch rows and columns. To obtain a unified attention distribution, we average over all answer queries and attention heads:
\begin{equation}
  \widehat{\mathbf{A}} = \frac{1}{N_A \, K} \sum_{n=1}^{N_A} \sum_{k=1}^{K} \tilde{\mathbf{A}}^{(\ell,k)} \in \mathbb{R}_+^{h\times w},
\end{equation}
and normalize across the spatial dimensions to yield the final importance distribution:
\begin{equation}
  \widetilde{\mathbf{A}} = \mathrm{Normalize}_{h,w}\bigl(\widehat{\mathbf{A}}\bigr).
\end{equation}

\section{Details of the Dataset}
\label{appendix:dataset}
\subsection{Caption Annotation}
As stated previously in Section~\ref{sec:vqa-verify}, the bounding-boxes used in our dataset are based on the annotations provided by \citet{shao2024visual} and \citet{wang2020general}. For image captioning, we utilized GPT-4o, a cutting-edge model in the field. After generating captions, we carried out manual review and refinement to ensure quality, and subsequently incorporated the VQA-Verify dataset into our work.
The following prompt was used to generate caption annotations for the dataset.

\begin{tcolorbox}[colback=gray!5!white,colframe=black!75!black,title=Prompt for Caption Annotation,breakable]
\begin{verbatim}
<image> Describe this image in
general.
\end{verbatim}
\end{tcolorbox}

\begin{figure*}[htbp]
  \centering

  \begin{subfigure}{\linewidth}
    \centering
    \includegraphics[width=\textwidth]{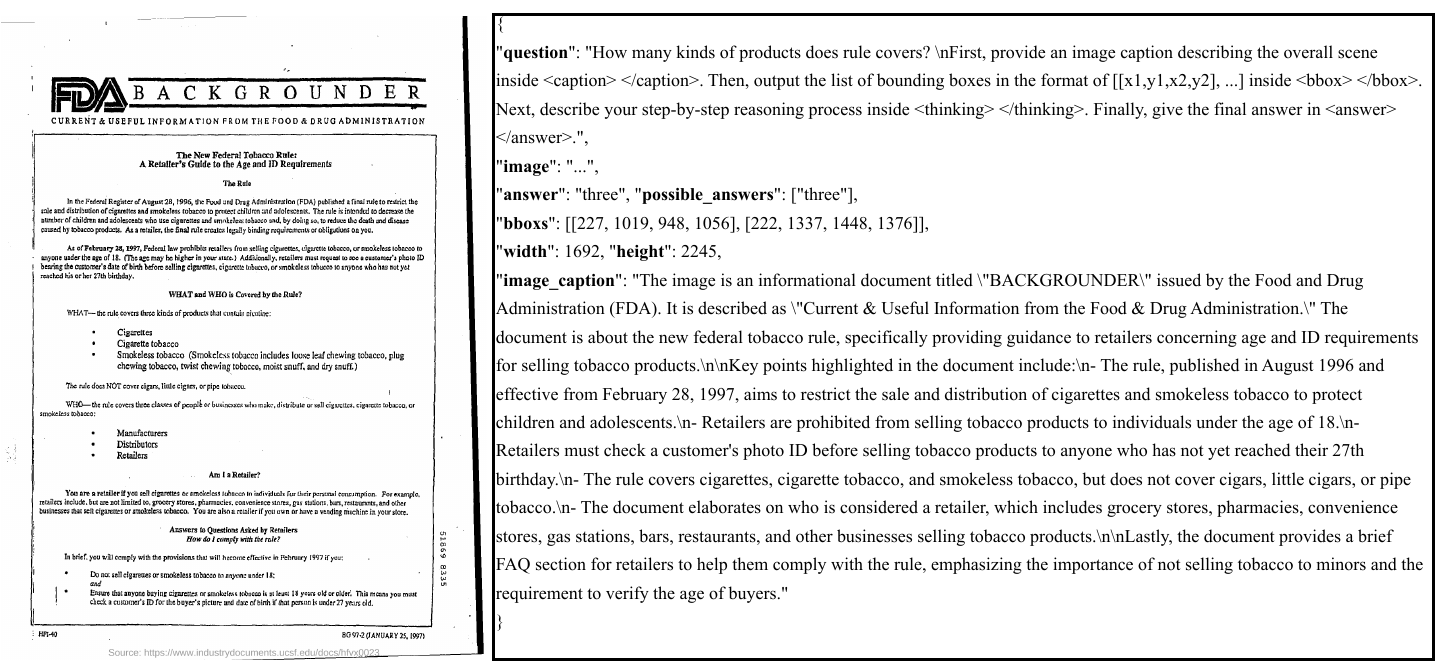}
  \end{subfigure}

  \begin{subfigure}{\linewidth}
    \centering
    \includegraphics[width=\textwidth]{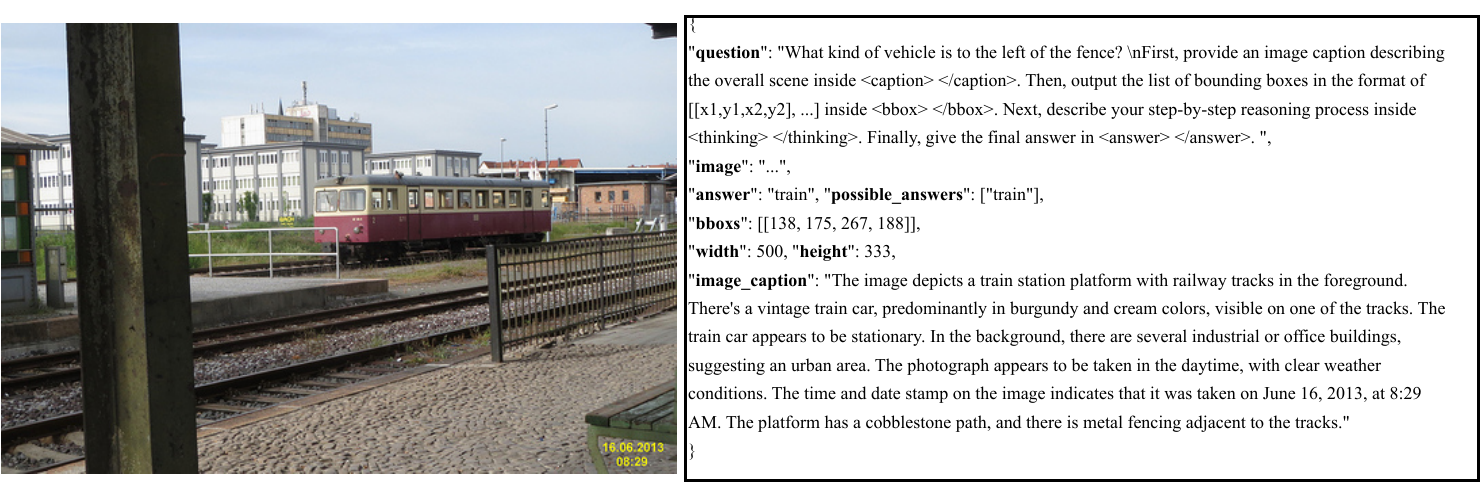}
  \end{subfigure}

  \caption{Examples of VQA-Verify.}
  \label{fig:dataset_example}
\end{figure*}

Examples of VQA-Verify are illustrated in Figure~\ref{fig:dataset_example}.


\subsection{Statics of VQA-Verify}

Table~\ref{tab:combined-stats} summarizes the overall size and annotation density of VQA-Verify, as well as the composition and average grounding signal per source dataset. The first row reports the total number of training samples, the average number of bounding-boxes per sample, and the average caption length in words. Subsequent rows break down these metrics for each of the integrated benchmark datasets, illustrating variation in visual complexity (boxes) and descriptive detail (caption length). The estvqa~\cite{wang2020general} dataset integrates data from Text-Total~\cite{ch2020total}, ICDAR2013~\cite{karatzas2013icdar}, ICDAR2015~\cite{karatzas2015icdar}, CTW1500~\cite{liu2019curved}, COCO-Text~\cite{veit2016coco}, LSVT~\cite{sun2019chinese}, and MLT~\cite{nayef2019icdar2019}. It is worth noting that the bounding-box annotations provided by \citeauthor{wang2020general} are defined by four points rather than the standard rectangular format. To ensure consistency, we converted them to the common \([x_1, y_1, x_2, y_2]\) representation by computing the enclosing rectangle.

\begin{table*}[h!]
\centering
\small
\begin{tabular}{lrrr}
\toprule
\textbf{Source} & \textbf{\# Samples} & \textbf{Avg BBoxes / Sample} & \textbf{Avg Caption Words} \\
\midrule
Train                            & 12,000               & 1.34                         & 112.12                     \\
\midrule
cub-200~\cite{wah2011caltech}                             & 1,000                & 1.00                         & 76.03                      \\
docvqa~\cite{mathew2021docvqa}                         & 1,000                & 2.17                         & 145.53                     \\
dude~\cite{van2023document}                            & 1,000                & 1.00                         & 150.27                     \\
estvqa~\cite{wang2020general}                         & 1,000                & 1.00                         & 98.53                      \\
gqa~\cite{hudson2019gqa}                               & 1,000                & 1.00                         & 86.53                      \\
infographicsvqa~\cite{mathew2022infographicvqa}        & 1,000                & 2.31                         & 200.11                     \\
openimages~\cite{kuznetsova2020open}                   & 1,000                & 1.00                         & 81.27                      \\
sroie~\cite{huang2019icdar2019}                        & 1,000                & 1.00                         & 146.31                     \\
textcap~\cite{sidorov2020textcaps}                     & 1,000                & 1.84                         & 96.93                      \\
textvqa~\cite{singh2019towards}                        & 1,000                & 1.72                         & 95.77                      \\
v7w~\cite{zhu2016visual7w}                              & 1,000                & 1.00                         & 87.08                      \\
vsr~\cite{liu2023visual}                               & 1,000                & 1.00                         & 80.97                      \\
\bottomrule
\end{tabular}
\caption{Overall and per-source statistics for VQA-Verify.}
\label{tab:combined-stats}
\end{table*}

\subsection{Dataset Verification}
\label{appendix:dataset_manual}

To ensure the quality of the automatically generated captions and bounding-boxes in VQA-Verify, we performed a manual verification on a sampled subset. Below we describe the verification procedure and summarize the results.

We randomly sampled 1,500 instances (12.5\% of the 12,000 total samples) and and conducted a manual review of each. During this review, captions were evaluated to ensure they accurately reflected key image content, contained between 10 and 20 words, and were free from spelling or semantic errors; bounding-boxes were verified to tightly enclose the region relevant to the answer, with an Intersection-over-Union (IoU) of at least 0.8 relative to the original automatic annotation; and answers were checked for consistency with both the question and the image. Results are shown in Table~\ref{tab:manual-verification}. 

\begin{table*}[h!]
\centering
\small
\begin{tabular}{lrrrr}
\toprule
\textbf{Item}                   & \textbf{Failures} & \textbf{Failure Rate} & \textbf{Common Issues}           \\
\midrule
Caption Quality Check                          & 27               & 1.8\%                 & Overly verbose, missing details  \\
bounding-box Accuracy Check                    & 18                & 1.2\%                 & Box misalignment, incomplete area \\
Answer Consistency Check                      & 9                & 0.6\%                 & Mismatch with image/question     \\
\bottomrule
\end{tabular}
\caption{Summary of manual verification results.}
\label{tab:manual-verification}
\end{table*}

\section{Variance Analysis}
\label{appendix:variance}

To rigorously characterize the sources of policy gradient variance in GRPO, we leverage the \emph{Law of Total Variance} to decouple the variance into intra‑trajectory and inter‑trajectory components. Let $\nabla J_{\mathrm{GRPO}}$ denote the policy gradient estimator:

\begin{equation}
\begin{split}
\nabla_{\theta}J_{\mathrm{GRPO}} & \approx
\mathbb{E}_{q,o\sim\pi_{\theta_{\mathrm{old}}}}\Biggl[
\frac{1}{G}\sum_{i=1}^{G}
\frac{1}{|o_{i}|} \\
& \quad \times \sum_{t=1}^{|o_{i}|}
h_{i,t}\,\nabla_{\theta}\log\pi_{\theta}(o_{i,t}\mid q,o_{i,<t})\;\tilde A_{i,t}
\Biggr]
\end{split}
\label{eq:gradient_estimation}
\end{equation}

In this expression, we omit the KL‐divergence penalty and clipping terms from PPO since they are deterministic functions of the gradient and do not contribute to sampling noise. The \emph{total variance} of the estimator can be decomposed as:

\begin{equation}
\begin{split}
\mathrm{Var}(\nabla J_{\mathrm{GRPO}})
=
\underbrace{\mathbb{E}_{\tau}\bigl[\mathrm{Var}(\nabla J_{\mathrm{GRPO}}\mid \tau)\bigr]}_{\text{Intra‑Trajectory Variance}} \\
\;+\;
\underbrace{\mathrm{Var}_{\tau}\bigl(\mathbb{E}[\nabla J_{\mathrm{GRPO}}\mid \tau]\bigr)}_{\text{Inter‑Trajectory Variance}}
\end{split}
\label{eq:decomposed}
\end{equation}

Here, $\tau$ denotes a sampled trajectory. The inter‑trajectory term reflects variance due to differences in total rewards $R(\tau)$ across trajectories. In GRPO, the advantage at each token is normalized by the group statistics:

\[
\tilde r_i
=
\frac{r_i - \mathrm{mean}(\mathbf r)}
     {\mathrm{std}(\mathbf r)},
\quad
\mathbf r = \{r_i\}_{i=1}^G.
\]

Thus the trajectory‑conditional expected gradient scales as

\begin{equation}
\mathbb{E}\bigl[\nabla J_{\mathrm{GRPO}}\mid \tau\bigr]
\;\propto\;
\frac{1}{|o|}
\sum_{t=1}^{|o|}
h_t \cdot \nabla_{\theta}\log\pi_{\theta}(o_{t}\mid q,o_{<t}) \cdot \tilde r,
\label{eq:gradient_cond_exp}
\end{equation}
\begin{equation}
g_{i,t} = \nabla_{\theta}\log\pi_{\theta}(o_{i,t}\mid q,o_{i,<t}).
\label{eq:g_definition}
\end{equation}

Since trajectories are sampled independently but token generation within each trajectory is autoregressive, we have:

\begin{equation}
\mathrm{Var}_{\tau}\bigl(\mathbb{E}[\nabla J_{\mathrm{GRPO}}\mid \tau]\bigr)
\;\propto\;
\mathrm{Var}\bigl(R(\tau)\bigr),
\quad
R(\tau) = \sum_{k=1}^n \beta_k\,R_k.
\end{equation}

This shows that reducing the variance of the total reward $R(\tau)$ directly lowers the inter‑trajectory variance and thus stabilizes gradient estimates.

Our observation in experiments shows that even though the verifiable reasoning patterns reward and the accuracy reward are often positively correlated, the overall variance of the total reward still decreases. This phenomenon can be explained by the \emph{diversification effect}~\cite{markowitz1952portfolio}: when the total reward is constructed as a weighted combination of multiple sub-rewards with weights \(\{\beta_k\}\), the overall variance can be reduced even if the components are positively correlated.

According to the variance formula for a weighted sum:
\[
\mathrm{Var}\bigl(R(\tau)\bigr)
=
\sum_{i}\beta_i^2\mathrm{Var}(R_i)
+ 2\sum_{i<j}\beta_i\beta_j\mathrm{Cov}(R_i, R_j),
\]
the total variance depends not only on the variances of the individual components but also on their pairwise covariances. Even when \(\mathrm{Cov}(R_i, R_j) > 0\), the squared weights \(\beta_i^2 < 1\) for all \(i\) dilute the contribution of each term, and as long as the correlation coefficients \(\rho_{ij} < 1\), the combined variance can be strictly smaller than the variance of a single reward term.

In the context of GRPO, the verifiable reasoning reward emphasizes consistency in intermediate reasoning steps (e.g., caption or bounding-box grounding), while the accuracy reward focuses on the final answer correctness. Although the two rewards are positively correlated, they capture complementary aspects of the task. By allocating appropriate weights, we retain useful signal from both while suppressing the noise associated with each individual component. This diversification not only reduces inter-trajectory variance but also leads to more stable gradient estimates and improved convergence behavior during training.


\section{Implementation Details}
\label{appendix:implementation}
Our experiments are conducted using Qwen2.5‑VL‑Instruct‑3B. The model’s \texttt{MAX\_PIXELS} is set to $512\times28\times28$, and \texttt{MIN\_PIXELS} to $256\times28\times28$. Training is performed on eight NVIDIA H100 Tensor Core GPUs. For the dataset, we utilize VQA‑Verify (see Section~\ref{sec:vqa-verify}), which enables the incorporation of intermediate caption and bounding‑box reward signals during training. In the reward configuration, we assign equal weight to all reward signals. That is, all values of $\beta$ are set to $1/k$.

We do not perform a cold‑start; instead, we train directly using GRPO. During training, we set \texttt{max\_length} to 2048, and the GRPO group size $G$ to 16, corresponding to a per‑device batch size of 4. The sampling parameters are configured as follows: \texttt{temperature} = 1.0, \texttt{top\_k} = 50, \texttt{top\_p} = 0.9, and \texttt{repetition\_penalty} = 1.0. We use a learning rate of $1\times10^{-6}$ and perform full fine‑tuning for one epoch. The clip range is set to 0.2, and the KL divergence coefficient to 0.05. Throughout training, only the linear layers are updated, while the visual encoder remains frozen. During training, we randomly selected 1\% of the training set as the validation set. The system prompt used in the GRPO training process is as follows:

\begin{tcolorbox}[colback=gray!5!white,colframe=black!75!black,title=System Prompt for GRPO training,breakable]
\begin{verbatim}
A conversation between User and
Assistant in a Visual Question 
Answering (VQA) task. The User asks
a question about an image, and the
Assistant solves it. Given an image
and a question, follow these steps:

First, generate a brief image
caption describing the overall
scene. Provide the caption inside
<caption> </caption>.

Next, identify the most relevant
image regions for answering the
question. Enclose these coordinates
in <bbox>[[x1,y1,x2,y2], ...]
</bbox>.

Then, formulate a step-by-step
thinking process that outlines the
reasoning required to arrive at the
solution. Enclose this reasoning in
<thinking> </thinking> tags.

Finally, provide the final answer to
the question inside <answer>
</answer> tags.
\end{verbatim}
\end{tcolorbox}

\section{More Experiments}
\label{appendix:more_experiments}
\subsection{Detailed Results on MMStar}
\begin{figure*}[htbp]
  \centering
 \includegraphics[width=0.7\textwidth]{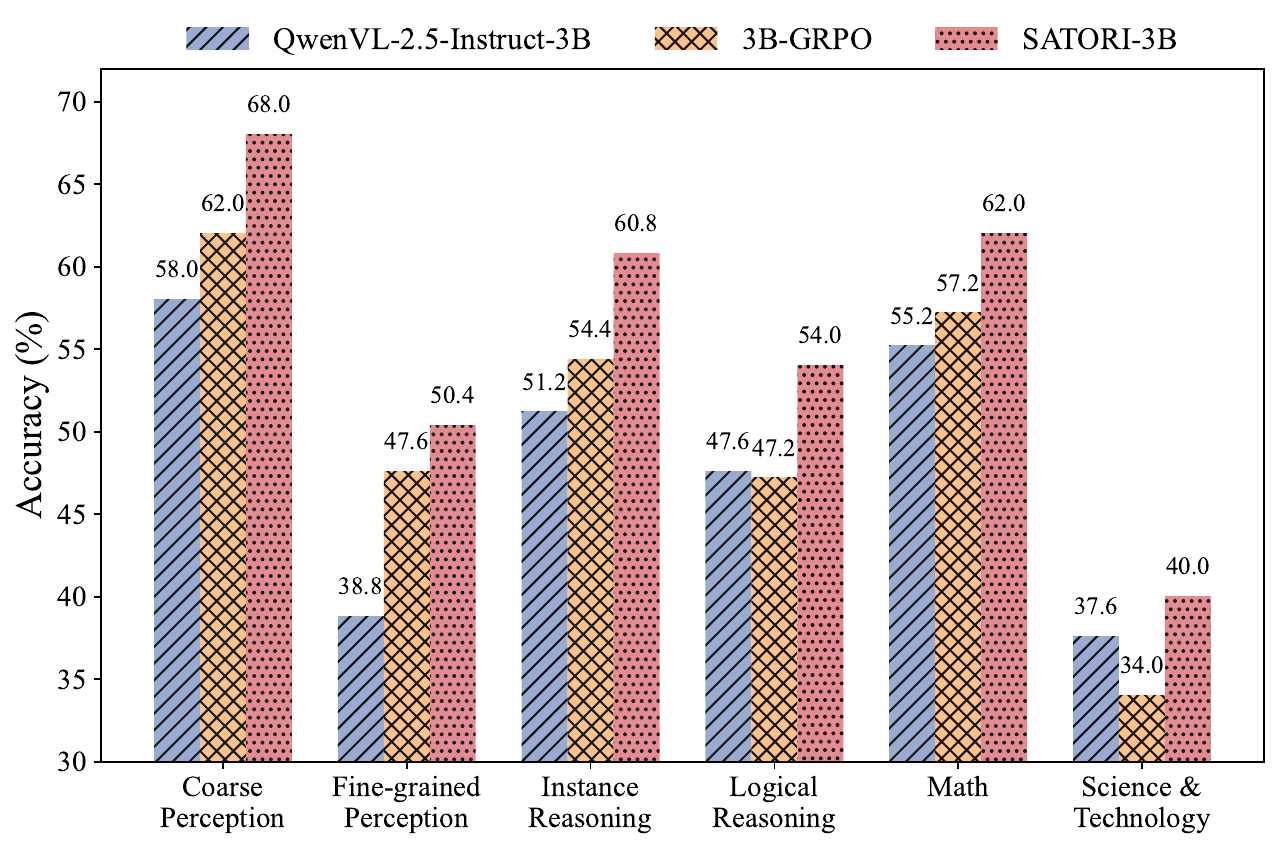}
  \caption{Comparison of model average performance on each category. SATORI outperforms the 3B-GRPO baseline, demonstrating the effectiveness of using verifiable reasoning patterns as rewards.}
  \label{fig:mmstar}
\end{figure*}

We present detailed comparative results of 3B-size models on the MMStar dataset in Figure~\ref{fig:mmstar}. The results show that our method consistently outperforms the 3B-GRPO baseline—which uses the same dataset and settings but lacks verifiable signals—across all categories. Notably, on more complex reasoning and math tasks, SATORI surpasses it by more than 5\%.

\begin{table*}[tb]
    \centering
     \caption{\textbf{Comparison of SATORI with other MLLMs and methods in MMBench~\cite{liu2024mmbench}.} SATORI outperforms other open-source models, surpasses alternative reasoning-based MLLM approaches, and achieves competitive performance across most benchmarks. Specifically, LR denotes Logical Reasoning, AR denotes Attribute Reasoning, RR denotes Relation Reasoning, PPR denotes Physical Property Reasoning, SITU represents Structuralized Image-Text Understanding, FP-C represents Fine-grained Perception (Cross Instance), FP-S represents Fine-grained Perception (Single Instance), and CP refers to Coarse Perception. Results marked with $\dagger$ are sourced from~\cite{duan2024vlmevalkit}.}
    \tiny
    \setlength\tabcolsep{2pt} 
    \renewcommand{\arraystretch}{1.1} 
    \scalebox{1.09}{
    \begin{tabular}{r |cccc |ccc |ccccc |ccc |cc |ccc |l}
     \multicolumn{1}{c}{ } &
     \multicolumn{4}{c}{FP-S} &
     \multicolumn{3}{c}{FP-C} &
     \multicolumn{5}{c}{CP} &
     \multicolumn{3}{c}{AR} &
     \multicolumn{2}{c}{LR} &
     \multicolumn{3}{c}{RR} &
       \\
      Model/Method &
      \rotatebox{90}{Action Recognition} &
      \rotatebox{90}{Object Localization}  &
      \rotatebox{90}{Celebrity Recognition} &
      \rotatebox{90}{OCR} &
      \rotatebox{90}{Spatial Relationship} &
      \rotatebox{90}{Attribute Comparison} &
      \rotatebox{90}{Attribute Recognition} &
      \rotatebox{90}{Image Emotion} &
      \rotatebox{90}{Image Quality} &
      \rotatebox{90}{Image Scene} &
      \rotatebox{90}{Image Style} &
      \rotatebox{90}{Image Topic} &
      \rotatebox{90}{Function Reasoning} &
      \rotatebox{90}{Identity Reasoning} &
      \rotatebox{90}{PPR} &
      \rotatebox{90}{Future Prediction} &
      \rotatebox{90}{SITU} &
      \rotatebox{90}{Nature Relation}  &
      \rotatebox{90}{Physical Relation} &
      \rotatebox{90}{Social Relation} & Avg. \\
      \hline
      
      GPT-4V$^\dagger$ & 73.5 & 36.2 & 61.2 & 93.3 & 44.0 & 46.7 & 78.7 & 56.7 & 37.9 & 81.9 & 76.1 & 94.4 & 88.9 & 96.1 & 53.2 & 65.3 & 56.0 & 68.5  & 33.3 & 64.8 & 65.4   \\
      
      Gemini-1.5 Pro$^\dagger$ & 85.5 & 69.5 & 84.4 & 77.8 & 66.7 & 65.3 & 93.3 & 74.4 & 49.2 & 84.7 & 79.3 & 75.6 & 85.6 & 94.7 & 64.6 & 61.3 & 65.1 & 83.7 & 52 & 69.2 & 74.6  \\
      Gemini-2.0 Flash$^\dagger$ & 86.3 & 66.7 & 72.1 & 74.4 & 65.3 & 78.7 & 70.8 & 64.4 & 54.0 & 74.3 & 65.2 & 78.9 & 80.0 & 78.9 & 60.8 & 64.0 & 63.3 & 79.3 & 50.7 & 75.8 & 70.4 \\
      Llava-Next-8B$^\dagger$ & 80.3 & 64.8 & 74.8 & 83.3 & 44.0 & 66.7 & 79.8 & 78.9 & 48.4 & 85.4 & 79.3 & 97.8 & 88.9 & 97.4 & 64.6 & 66.7 & 37.6 & 66.3 & 50.7 & 84.6 & 72.1 \\
      Llava-CoT-11B$^\dagger$ & 81.2 & 62.9 & 89.1 & 92.2 & 48 & 62.7 & 86.5 & 74.4 & 46.8 & 83.3 & 75.0 & 88.9 & 90.0 & 98.7 & 67.1 & 66.7 & 49.5 & 79.3 & 56.0 & 82.4 & 74.4 \\
       \hline
       \rowcolor{gray!10}
    \multicolumn{1}{l|}{\textit{Qwen2.5-VL-Ins-3B}} &  &  &  &  &  &  &  &  &  &  &  &  &  &  &  &  &  &  &  & &
    \\
        Original & 78.2 & 40.0 & 66.7 & 70.0 & 22.0 & 29.4 & 56.7 & 81.7 & 21.4 & 89.6 & 59.7 & 76.7 & 83.3 & 96.1 & 41.5 & 42.0 & 45.9 & 66.1 & 40.0 & 88.7 & 60.8 \\ 
        GRPO-3B & 76.9 & 52.9 & 94.9 & 75.0 & 38.0 & 27.5 & \textbf{86.7} & 81.7 & \textbf{39.3} & 56.2 & \textbf{83.9} & 43.3 & 86.7 & \textbf{98.0} & 34.0 & 42.0 & 56.8 & 56.5 & 56.0 & 85.5 & 64.6 \\ 
   \rowcolor{green!10} SATORI-3B w/o thinking & \textbf{83.3} & \textbf{60.0} & \textbf{97.0} & \textbf{91.7} & \textbf{46.0} & \textbf{66.7} & \textbf{90.0} & \textbf{88.3} & 38.1 & \textbf{93.8} & 82.3 & \textbf{91.7} & \textbf{86.7} & \textbf{98.0} & \textbf{52.8} & \textbf{54.0} & \textbf{58.1} & \textbf{80.6} & \textbf{62.0} & \textbf{91.9} & \textbf{76.5} \\
    \end{tabular}
}
\captionsetup{font={small}}
\label{tab:mmbench}
\end{table*}

\subsection{Detailed Results on MMBench}
As shown in Table \ref{tab:mmbench}, our SATORI-3B w/o thinking achieves an average score of 76.5\%. This significantly outperforms both the Original baseline (60.8\%) and the R1-like GRPO-3B (64.6\%), demonstrating the superiority of the SATORI framework.

SATORI's 76.5\% score is highly competitive, surpassing other strong open-source models like Llava-Next-8B (72.1\%) and Llava-CoT-11B (74.4\%). Notably, it also exceeds the performance of closed-source models, including Gemini-1.5 Pro (74.6\%) and GPT-4V (65.4\%).

Analyzing the sub-tasks reveals SATORI's dominance, where it achieves the top score in 16 out of 20 categories compared to the GRPO baseline. The most significant gains are in Fine-grained Perception (FP-S \& FP-C), such as in OCR (91.7 vs. 75.0) and Attribute Comparison (66.7 vs. 27.5). This confirms our hypothesis that SATORI's spatial anchoring effectively solves the "attention dilution" problem inherent in free-form reasoning methods like GRPO. The broad improvements across categories like Relational Reasoning (RR) further validate SATORI's effectiveness and strong generalization.

\subsection{Sensitivity Analysis of Reward Weights}
\label{sec:sensitivity_analysis}

In our default configuration, we assign equal weights to the three reward signals (i.e., $\mathcal{R}_{caption}$, $\mathcal{R}_{bbox}$, and $\mathcal{R}_{ans}$ are each weighted at $1/3$). To verify the robustness of SATORI and ensure that our performance gains are not derived from sensitive hyperparameter tuning, we conducted a sensitivity analysis by varying these weights.

We evaluated several weight distributions on the MMStar benchmark using the Qwen2.5-VL-Instruct-3B backbone and the non-thinking version. As shown in Table~\ref{tab:reward_weights}, the model demonstrates high stability across different configurations. Shifting the emphasis toward the final answer (e.g., $[1/4, 1/4, 1/2]$) or the intermediate caption (e.g., $[1/2, 1/4, 1/4]$) results in only minor performance fluctuations, with accuracy ranging from 55.1\% to 56.0\%. These results confirm that SATORI is not overly sensitive to specific reward weights and achieves consistent improvements without requiring extensive hyperparameter search.

\begin{table}[h]
    \centering
    \caption{Sensitivity analysis of reward weights on the MMStar benchmark. The results demonstrate that the model's performance remains robust across different weight distributions for caption, bounding box, and answer rewards.}
    \label{tab:reward_weights}
    \begin{tabular}{lc}
        \toprule
        Weight Config ($\mathcal{R}_{cap}, \mathcal{R}_{bbox}, \mathcal{R}_{ans}$) & Accuracy  \\
        \midrule
        $[1/3, 1/3, 1/3]$ & 55.9 \\
        $[1/2, 1/4, 1/4]$ & 55.6 \\
        $[1/4, 1/4, 1/2]$ & 56.0 \\
        $[1/4, 1/2, 1/4]$ & 55.2 \\
        $[1/5, 1/5, 3/5]$ & 55.1 \\
        \bottomrule
    \end{tabular}
\end{table}

\subsection{Analysis of Attention Concentration}

To validate that each component of the SATORI framework (Caption, Bbox, and Think) contributes to better visual grounding, we conducted a detailed ablation study. This analysis expands on the simple comparison in Table~\ref{tab:ablation} by isolating the impact of each reward signal on the model's attention. All experiments use the Qwen2.5VL-3B-Instruct as the starting point. We performed inference on 1,000 samples randomly selected from the OpenImages dataset. This dataset was not part of our VQA-Verify training data, ensuring the models were evaluated on their generalization capabilities. We measured two key metrics: (1) Region Attention Density (RAD), our proposed metric to quantify attention concentration on answer-relevant regions, and (2) Accuracy, the final answer accuracy on the VQA tasks. To ensure a fair and direct comparison of reasoning-specific focus, we calculated the RAD only on the attention maps generated during the production of the final answer tokens. This approach isolates the model's visual grounding at the moment of decision-making, providing a clean comparison across all configurations.

The results of our attention ablation study are presented in Table~\ref{tab:rad_ablation_appendix}. The result confirms our "attention dilution" hypothesis. We observe a clear distinction between SFT and RL; while \texttt{+BBox+Caption+SFT} improved RAD to 0.3620, the equivalent \texttt{+BBox+Caption+RL} model was far more effective, achieving 0.4410 RAD. This suggests RL, unlike SFT, optimizes for grounding rather than just mimicking format. The \texttt{+BBox+RL} reward provided the single largest boost in focus (0.4120 RAD), proving it is the key driver of our method. The components are complementary, as the full \texttt{SATORI} model achieved the highest RAD (0.4588) and accuracy (88.5\%). This demonstrates that our verifiable rewards directly teach attention concentration, which correlates strongly with accuracy.

\begin{table}[thb] 
\centering 
\small
\caption{Component-wise ablation study on 1,000 unseen OpenImages samples. All RAD calculations are normalized by comparing attention \textit{only} during the generation of \texttt{<answer>} tokens to ensure fairness. The results show that SATORI's RL components progressively increase attention concentration, in sharp contrast to the attention dilution caused by free-form reasoning.} 
\label{tab:rad_ablation_appendix} 
    \begin{tabular}{l c c}
    \toprule 
    \textbf{Model Configuration} & \textbf{Avg. RAD} & \textbf{Accuracy (\%)} \\ 
    \midrule 
    \rowcolor{gray!10} Qwen2.5-VL-Ins-3B & 0.3029 & 79.0 \\ 
    \midrule 
    +Free Form Reasoning+RL & 0.3110 & 79.2 \\ 
    \midrule 
    +BBox+SFT & 0.3450 & 81.5 \\
    +BBox+Caption+SFT & 0.3620 & 82.8 \\
    \midrule 
    +Caption+RL & 0.3550 & 82.0 \\
    +Caption+Think+RL & 0.3670 & 83.1 \\
    +BBox+RL & 0.4120 & 85.5 \\
    +BBox+Think+RL & 0.4315 & 86.8 \\
    +BBox+Caption+RL & 0.4410 & 87.9 \\
    \textbf{SATORI (Full)} & \textbf{0.4588} & \textbf{88.5} \\
    \bottomrule 
    \end{tabular} 
\end{table}

\begin{table}[thb]
\centering
\small
\caption{Ablation study on the reasoning sequence permutations using the SATORI-7B model on MMStar. We compare all orderings of the \texttt{Caption} (Cap), \texttt{BBox} (Box), and \texttt{Think-Answer} (Think) components. The results validate our hypothesis that establishing full visual grounding (\emph{Grounding-First}) before logical deduction yields the highest accuracy. \emph{Think-First} variants offer a strong performance/efficiency trade-off, while \emph{Mixed-Order} grounding is suboptimal.}
\label{tab:sequence_permutation_ablation}
\begin{tabular}{l l c}
\toprule
\textbf{Strategy} & \textbf{Method} & {\textbf{Acc.} $\uparrow$}  \\
\midrule
Baseline & Qwen2.5-VL-7B & 64.1 \\ 
\midrule

\rowcolor{gray!10}
\textbf{Ground-First} & Cap $\rightarrow$ Box $\rightarrow$ Think & \textbf{69.5}  \\
 & Box $\rightarrow$ Cap $\rightarrow$ Think & 69.2 \\
\midrule

Think-First & Think $\rightarrow$ Cap $\rightarrow$ Box & 65.5  \\
 & Think $\rightarrow$ Box $\rightarrow$ Cap & 65.4 \\
\midrule

Mixed-Order & Box $\rightarrow$ Think $\rightarrow$ Cap & 64.8  \\
 & Cap $\rightarrow$ Think $\rightarrow$ Box & 64.5 \\
\bottomrule
\end{tabular}
\end{table}

\subsection{Ablation Study on Caption Reward Function}
\label{appendix:reward_ablation}

To validate the choice of our caption reward $\mathcal{R}_{caption}$ and address the critique of lexical-overlap metrics (e.g., BLEU, ROUGE) as poor semantic proxies, we conducted a detailed ablation study. We replaced our standard lexical-overlap metric with several state-of-the-art semantic similarity measures, including bi-encoder similarity (from \texttt{sentence-transformers}), token-level similarity (\texttt{BERTScore}), and cross-encoder similarity. Our experiments are conducted on Qwen-2.5-VL-Instruct-3B, with SATORI-3B w/o thinking as the baseline for comparison. All other experimental settings, including the hyperparameters, were held consistent with the main experiment setup. The results, presented in Table \ref{tab:caption_reward}, demonstrate that our original combination of BLEU-4 and ROUGE-L yields the best overall performance. While semantic metrics are theoretically more robust, they appear to provide a less stable or less direct reward signal for this specific task compared to the well-behaved, fast-to-compute lexical metrics.

\begin{table*}[th]
\centering
\caption{Ablation study on the $\mathcal{R}_{caption}$ reward function. Performance is compared between our standard lexical-overlap metric and several semantic similarity alternatives. All models are trained on the SATORI-3B setting.}
\label{tab:caption_reward}
\begin{tabular}{l l c c}
\toprule
\textbf{$\mathcal{R}_{caption}$ Setting} & \textbf{Metric / Model Used} & \textbf{MMBench} $\uparrow$ & \textbf{MMStar} $\uparrow$ \\
\midrule
\rowcolor{gray!10}
\textbf{Ours (Lexical)} & \textbf{BLEU-4 + ROUGE-L} & \textbf{76.5} & \textbf{55.9} \\
Lexical & BLEU-4 (only) & 75.8 & 53.5 \\
Lexical & ROUGE-L (only) & 76.1 & 54.6 \\
\midrule
Semantic (Bi-Encoder) & all-mpnet-base-v2 & 74.0 & 52.0 \\
Semantic (Token-level) & roberta-large (BERTScore) & 71.1 & 50.8 \\
Semantic (Cross-Encoder) & cross-encoder/stsb-roberta-large & 72.2 & 51.6 \\
\bottomrule
\end{tabular}
\end{table*}

\subsection{Ablation Study on Reasoning Sequence Permutations}
\label{sec:appendix_ablation_sequence}

To validate our \textbf{Glance $\rightarrow$ Focus $\rightarrow$ Think} design, we conducted an ablation study on all $3! = 6$ sequence permutations of the \texttt{Caption} (Cap), \texttt{BBox} (Box), and \texttt{Think-Answer} (Think) components. As shown in Table~\ref{tab:sequence_permutation_ablation}, we evaluated all SATORI-7B variants (Qwen2.5-VL-7B backbone) on the MMStar benchmark. The empirical results strongly support our hypothesis: the \emph{Ground-First} strategies, which establish grounding *before* reasoning, significantly outperform all other configurations. Our full SATORI paradigm (\texttt{Cap $\rightarrow$ Box $\rightarrow$ Think}) achieves the peak accuracy of 69.5\%, slightly outperforming the \texttt{Box $\rightarrow$ Cap $\rightarrow$ Think} variant (69.2\%). This confirms our "Glance-then-Focus" approach as the optimal design.

Conversely, the \emph{Think-First} variants show a substantial performance drop (approx. -4 points), demonstrating that answering *before* grounding limits reasoning capability. The \emph{Mixed-Order} strategies perform worst, barely improving over the baseline (64.1\%) and confirming that interleaving these steps is detrimental. This permutation analysis confirms that the \texttt{Glance $\rightarrow$ Focus $\rightarrow$ Think} sequence is not arbitrary, but the optimal structure for maximizing accuracy by ensuring logical deduction is fully conditioned on verified visual grounding.




\begin{table*}[h]
\centering
\caption{Causal intervention analysis on the 1,000-sample OpenImages test set. }
\label{tab:causal_intervention}
\begin{tabular}{llcc}
\toprule
\textbf{Model Configuration} & \textbf{Intervention Type} & \textbf{Avg. RAD} & \textbf{Accuracy (\%)} \\
\midrule
Baseline (Qwen2.5-VL-Ins-3B) & None & 0.3029 & 79.0 \\
SATORI (Full) & None & 0.4588 & 88.5 \\
\midrule
Baseline + Oracle-Focus & Explicit BBox (Text) & $\sim$1.0 (Imputed) & \textbf{90.2} \\
SATORI + Ablated-Focus & Explicit BBox (Text) & $\sim$0.0 (Imputed) & \textbf{31.5} \\
SATORI + Ablation Mask & \textbf{Implicit Attention} & $\sim$0.0 (Forced) & \textbf{28.2} \\
SATORI + Oracle Mask & \textbf{Implicit Attention} & $\sim$1.0 (Forced) & 83.9 \\
\bottomrule
\end{tabular}
\end{table*}

\section{Causal Analysis of Visual Focus and Accuracy}
\label{sec:causal_analysis}

A key hypothesis of this work is that SATORI's performance gains are \textit{causally} driven by its ability to mitigate the "visual-attention deficiency" (or "attention dilution") observed in free-form reasoning. Our main results (e.g., Figure 2) demonstrate a strong \textbf{correlation} between our Region Attention Density (RAD) metric and final accuracy.

However, to address the valid critique that this link is correlational, we present a series of interventional experiments to establish a \textbf{causal} link between focused visual grounding and model accuracy.

\subsection{Existing Causal Evidence in Prior Work}
Our work builds on established findings that have already demonstrated this causal link. For instance, the work of \citeauthor{zhang2025mllms} conducted a direct "interventional study" on baseline MLLMs. They manually forced the model's focus by providing "human-CROP" images that contained \textit{only} the ground-truth bounding box area. This intervention was shown to \textit{causally} and significantly improve performance, proving that the baseline model's limitation "stemmed from the model's inability to focus adequately".

Our contribution, therefore, is not in re-proving this fundamental principle, but in demonstrating a novel RL framework (SATORI) that can \textit{efficiently train} a model to \textit{learn} this focus mechanism on its own, replacing a "human-oracle" intervention with a verifiable, model-generated one.

\subsection{Interventional Experiments on SATORI}
To validate this causal mechanism \textit{within} our own framework, we conduct two new sets of experiments on the 1,000-sample OpenImages test set. We analyze interventions on both the \textbf{explicit BBox text} (the \texttt{Focus} output) and the \textbf{implicit visual attention} (the internal mechanism).

\begin{itemize}
    \item \textbf{Explicit BBox Intervention:}
    \begin{enumerate}
        \item \textbf{Baseline + Oracle-Focus (Text):} We take the original Qwen2.5-VL-Ins-3B baseline and \textit{feed it the ground-truth bounding box} in the prompt (\ie, "Given the region [x1, y1, x2, y2], answer the question.").
        \item \textbf{SATORI + Ablated-Focus (Text):} We take our full SATORI-3B model, let it generate its \texttt{<caption>} and \texttt{<bbox>}, but then \textit{replace} its predicted \texttt{<bbox>} text with an incorrect, random bounding box before it proceeds to the \texttt{Think} step.
    \end{enumerate}
    
    \item \textbf{Implicit Attention Intervention:}
    This is a more rigorous experiment that manipulates the model's internal state. We let the full SATORI-3B model generate its \texttt{Focus} (\texttt{<bbox>}) output, then map those text coordinates to the corresponding set of visual patches ($P_{bbox}$). We then apply an attention mask to the visual K-V cache \textit{only} for the subsequent \texttt{Think} and \texttt{Answer} generation steps.
    \begin{enumerate}
        \item \textbf{SATORI + Ablation Mask (Attention):} We \textit{prevent} the model from attending to the region it just identified. Attention scores for all visual patches \textit{inside} $P_{bbox}$ are set to 0.
        \item \textbf{SATORI + Oracle Mask (Attention):} We \textit{force} the model to \textit{only} attend to the region it identified. Attention scores for all visual patches \textit{outside} $P_{bbox}$ are set to 0.
    \end{enumerate}
\end{itemize}

\subsection{Results and Causal Analysis}
The results presented in Table \ref{tab:causal_intervention} provide decisive causal evidence for the efficacy of the SATORI framework. Specifically, providing the baseline model with "Oracle" focus boosts accuracy from 79.0\% to 90.2\%, confirming that visual attention deficiency is the primary bottleneck. Conversely, ablating SATORI's focus—either by providing incorrect bounding box text or by blinding the model to the corresponding visual patches causes performance to collapse to 31.5\% and 28.2\% respectively, while restricting attention solely to the predicted region maintains high accuracy (83.9\%).

Collectively, these findings move beyond simple correlation to establish a strong causal link. They demonstrate that SATORI's \texttt{Think} step is functionally and causally conditioned on the visual evidence identified during the \texttt{Focus} step. This confirms that the structured \texttt{Glance-Focus-Think} paradigm successfully enforces a necessary dependency on visual grounding for accurate reasoning, rather than merely acting as a formatting constraint.

\section{Discussion}
\subsection{Limitations}
\label{appendix:limit}
\textbf{Dependence on Base Model Instruction‑Following and Grounding.} While SATORI demonstrates significant benefits in zero-shot visual reasoning by leveraging a no–cold‑start GRPO training scheme atop powerful base MLLMs, several limitations and avenues for future exploration remain.
Our approach capitalizes on the strong instruction‑following and visual grounding abilities of models such as Qwen2.5‑VL. The ability to output bounding-boxes as intermediate rewards is a direct consequence of this pre‑training on grounding tasks. However, for weaker base models that lack such capabilities, a purely zero–cold‑start strategy may struggle. In these cases, an initial supervised fine‑tuning (SFT) phase with task‑specific data would likely be necessary to bootstrap both instruction adherence and structured reasoning. Similarly, models not pre‑trained on visual grounding would benefit from a phase of visual‑instruction tuning, exposing them to paired image, instruction, and bounding‑box annotations, before applying our reinforcement framework.

\subsection{Future Works}
\textbf{Towards Fine‑Grained, Step‑by‑Step Verification.} Our future work will explore a more fine‑grained verification framework in which, at each reasoning step, the model attends to and is rewarded on a distinct image region. By leveraging dynamic visual attention maps rather than a single bounding-box, we can decompose complex, multi‑step problems, particularly in mathematics, into a sequence of visually grounded subtasks. 

\textbf{Adaptive Stage Decomposition and Model‑Learned Structuring.}
Another promising direction is to move beyond a fixed four‑stage pipeline toward models that learn their own optimal decomposition of tasks. By introducing a learnable stage controller, SATORI could adapt the number and nature of intermediate steps to each question’s complexity. Meta‑learning or conditional computation techniques may enable the model to decide, at inference time, how many reasoning sub‑tasks are required and what form each should take (e.g., object detection, relation extraction, sub‑captioning).
